\documentclass{article}

\PassOptionsToPackage{numbers}{natbib}
\usepackage[main,final]{neurips_2026}

\usepackage{amsmath,amsfonts,bm}

\def\eqref#1{equation~\ref{#1}}

\def\1{\bm{1}}

\def\eps{{\epsilon}}

\def\vone{{\bm{1}}}
\def\vmu{{\bm{\mu}}}

\def\va{{\bm{a}}}

\def\vc{{\bm{c}}}

\def\ve{{\bm{e}}}

\def\vh{{\bm{h}}}

\def\vm{{\bm{m}}}

\def\vp{{\bm{p}}}
\def\vq{{\bm{q}}}
\def\vr{{\bm{r}}}

\def\vt{{\bm{t}}}
\def\vu{{\bm{u}}}
\def\vv{{\bm{v}}}

\def\vx{{\bm{x}}}
\def\vy{{\bm{y}}}
\def\vz{{\bm{z}}}

\def\mA{{\bm{A}}}

\def\mC{{\bm{C}}}
\def\mD{{\bm{D}}}
\def\mE{{\bm{E}}}

\def\mI{{\bm{I}}}

\def\mM{{\bm{M}}}

\def\mR{{\bm{R}}}
\def\mS{{\bm{S}}}

\def\mU{{\bm{U}}}

\def\mW{{\bm{W}}}
\def\mX{{\bm{X}}}

\def\mLambda{{\bm{\Lambda}}}

\DeclareMathAlphabet{\mathsfit}{\encodingdefault}{\sfdefault}{m}{sl}
\SetMathAlphabet{\mathsfit}{bold}{\encodingdefault}{\sfdefault}{bx}{n}

\usepackage[utf8]{inputenc} 
\usepackage[T1]{fontenc}    
\usepackage[colorlinks=true,citecolor=blue]{hyperref}       
\usepackage{url}            
\newcommand{\coderepourl}{\url{https://github.com/dsaragih/EAMS/tree/main}}
\usepackage{booktabs}       
\usepackage{amsfonts}       
\usepackage{nicefrac}       
\usepackage{microtype}      
\usepackage{xcolor}         
\usepackage{colortbl}       
\usepackage{lipsum}
\usepackage[breakable, theorems, skins]{tcolorbox}
\usepackage{scalerel}
\usepackage{stackengine,wasysym}

\newcommand{\pmval}[2]{#1\,{\scriptsize$\pm$\,#2}}

\usepackage{amsthm}
\usepackage{graphicx}
\usepackage{subcaption}
\usepackage{adjustbox}
\usepackage{dblfloatfix}
\usepackage{placeins}
\usepackage{wrapfig}
\usepackage{thmtools}
\usepackage{algorithm}
\usepackage{algpseudocode}

\newcommand{\tablestyle}[2]{\setlength{\tabcolsep}{#1}\renewcommand{\arraystretch}{#2}\centering\footnotesize}
\newcommand{\qualimgintra}[2][]{\includegraphics[width=\linewidth,trim=8cm 8cm 8cm 8cm,clip,#1]{#2_0.jpg}}
\newcommand{\qualimgteeth}[2][]{\includegraphics[width=\linewidth,trim=8cm 6cm 8cm 6cm,clip,#1]{#2_0.jpg}}
\newcommand{\qualimgliver}[2][]{\includegraphics[width=\linewidth,trim=8cm 7cm 8cm 7cm,clip,#1]{#2_0.jpg}}
\graphicspath{{./figures_jpg/}}

\title{Augmented Equivariant Mesh Networks for\\ Anatomical Segmentation}

\author{%
  Daniel Saragih\thanks{Work done during internship at Queen's University, Vector Institute} \\
  Department of Computing\\
  National University of Singapore\\
  \texttt{danielg.saragih@gmail.com} \\
}

\begin{document}

\maketitle

\begin{abstract}
Anatomical mesh segmentation requires models that operate directly on irregular surface geometry while remaining robust to changes in coordinate pose across meshes of varying resolution. Existing task-specific mesh and point-cloud methods are not equivariant, and can degrade sharply under test-time perturbation, for example dropping by 28--30 IoU points on 3D-IOSSeg at 40\textdegree{} rotation even with matched training augmentation. We present EAMS, an \textbf{E}quivariant \textbf{A}natomical \textbf{M}esh \textbf{S}egmentor built on Equivariant Mesh Neural Networks (EMNN), and evaluate it in four dataset settings across three clinical application areas, spanning edge-, vertex-, and face-level supervision. We combine intrinsic mesh descriptors with anatomy-aware priors, including PCA-derived frames for dental arches and liver surfaces, and augment message passing to provide lightweight global context. Across intracranial aneurysm and intraoral segmentation, EAMS variants are competitive with specialized baselines on unperturbed inputs while remaining stable under geometric perturbations, and on liver surfaces they expose a favorable trade-off between canonical-pose accuracy and rotation robustness. These results show that a lightweight ($<2\text{M}$ parameters) equivariant framework can deliver robust anatomical mesh segmentation across diverse label types.
\end{abstract}

\section{Introduction}
\label{sec:intro}
Anatomical mesh segmentation increasingly relies on 3D surface meshes rather than volumetric grids \cite{martin2015orthodontic,koo2017deformable,yang2020intra}: meshes are compact, topology-aware representations and the native output of many clinical reconstruction pipelines.
Tasks such as liver-surface labelling, intraoral tooth segmentation, and intracranial aneurysm delineation require methods that can operate directly on triangle-mesh geometry with varying resolution, noise, and pose.

\textbf{Challenges.}
Effective mesh segmentation must satisfy two competing demands.
First, predictions should be invariant to rigid-body transformations: arbitrary coordinate pose should not change the anatomical labels. Consistent acquisition alignment can nevertheless carry useful predictive information, motivating a distinction between coordinate pose and dataset-specific anatomical priors.
Second, mesh resolution in medical datasets varies dramatically (e.g.\ 3{,}000--20{,}000 edges for liver surfaces), requiring feature representations that are robust to sampling density while still capturing local geometry.
Existing point-cloud methods~\citep{qi2017pointnet++,wang2019dynamic,wu2019pointconv,wu2024point} are natively resolution-agnostic but do not exploit the topological structure of meshes; current segmentation methods that operate on meshes~\citep{yang2020intra,ben20233dteethseg,li2024fine,xi20253d,zhang2025nested} are not invariant and thus vulnerable to spurious correlations with pose.

\textbf{This work.}
We address these challenges with EAMS, an anatomical mesh segmentor built on Equivariant Mesh Neural Networks (EMNN)~\citep{trang20243}.\footnote{Code is publicly available at \coderepourl.}
EAMS combines intrinsic shape descriptors, anatomy-relative coordinates, and multi-token global context in a lightweight shared encoder for edge-, vertex-, and face-level segmentation. 
Our contributions are:
\begin{itemize}
    \item \textbf{Controlled anatomical mesh evaluation.} Across four dataset settings in three application areas, we evaluate a shared encoder under edge-, vertex-, and face-level supervision. Matched orientation augmentation reveals stronger perturbation robustness, competitive IntrA and dental accuracy, and a canonical-accuracy rotation-robustness trade-off on liver.
    \item \textbf{Anatomy-aware featurisation.} We combine intrinsic geometry with PCA-derived anatomical coordinates, state their frame-selection conditions, and test their rotation sensitivity. A raw-coordinate diagnostic distinguishes useful acquisition alignment from arbitrary pose.
    \item \textbf{Multi-token global context.} Beyond EMNN's shared global node, we study soft regional aggregators as a structured compression bottleneck and virtual nodes as a global memory bank. Single-fold ablations show gains on IntrA and 3D-IOSSeg that widening EMNN alone does not reproduce.
    
\end{itemize}

\section{Background and Related Work}
\label{sec:background}
\paragraph{Triangle meshes}
A triangle mesh is a tuple $\mathcal{M} = (\mathcal{V}, \mathcal{E}, \mathcal{F})$, where $\mathcal{V} = \{1, \ldots, n\}$ indexes the vertices, $\mathcal{E} \subseteq \mathcal{V} \times \mathcal{V}$ is the undirected edge set, and $\mathcal{F} \subseteq \mathcal{V}^3$ is the set of triangular faces, each an ordered triplet of vertices.
Each vertex $i \in \mathcal{V}$ carries a coordinate $x_i \in \mathbb{R}^3$.
A valid triangle mesh is a manifold: every edge belongs to at most two faces, no two distinct faces intersect except along a shared edge or vertex, and the faces around every vertex form a topological disk.
We write $N(i) = \{j : (i, j) \in \mathcal{E}\}$ for the one-ring neighbourhood of vertex $i$ and $\tau(i) = \{(j, k) : (i, j, k) \in \mathcal{F}\}$ for its adjacent faces.

\paragraph{Geometric features}
Every face $(i, j, k) \in \mathcal{F}$ induces a normal vector $n_{ijk} = (x_j - x_i) \times (x_k - x_i) \in \mathbb{R}^3$, whose magnitude equals twice the face area: $a_{ijk} = \|n_{ijk}\| / 2$.
Vertex-level normals can be obtained by aggregating over adjacent faces; an area-weighted average gives $n_i = \sum_{(j,k) \in \tau(i)} a_{ijk}n_{ijk} \,/\, \|\sum_{(j,k) \in \tau(i)} a_{ijk} n_{ijk}\|$.
These quantities play a central role in our model: the cross product $(x_j - x_i) \times (x_k - x_i)$ appears directly in the surface-aware face messages (Eq.~\ref{eq:mijk-emnn}) and in the equivariant coordinate update (Eq.~\ref{eq:xi-update}), encoding face geometry in a way that is invariant to translation and equivariant to rotation.

\paragraph{Deep learning on meshes}
Working with mesh data requires methods that incorporates the graph topology and geometric features of the mesh. Equivariant methods commonly rely on the invariants of their input features, which are then updated with learnable functions: prominent examples include E(n)-equivariant graph neural networks \cite{satorras2021n} and their extension to meshes, EMNN \cite{trang20243}, and simplices \citep{eijkelboom2023n}. EMNN has the upside of simplicity, as opposed to more complex convolutional \citep{lim2018simple, hanocka2019meshcnn,gong2019spiralnet++,feng2019meshnet,de2020gauge} and attention-based methods \citep{basu2022equivariant}.

\paragraph{Medical mesh segmentation}
These span numerous clinically relevant applications, including liver surface segmentation \cite{zhang2025nested}, intracranial aneurysm segmentation \cite{yang2020intra}, and intraoral scan tooth segmentation \cite{ben20233dteethseg, li2024fine}. We focus on these three tasks as they involve labels on each of the components of the mesh (edges, vertices, and faces, respectively), allowing us to evaluate the versatility of our method across different segmentation paradigms.

\section{Methods}
\label{sec:methods}
\subsection{Datasets}
\textbf{Liver surfaces. }
We retrieve a dataset of 200 manually annotated liver surface meshes \citep{zhang2025nested} drawn from three public CT datasets. The task is edge-level segmentation of two anatomical regions, the falciform ligament and the liver ridge, reviewed by clinical experts. Mesh resolution varies substantially (3{,}000--20{,}000 edges) due to natural morphological variation across subjects, posing a primary robustness challenge for equivariant segmentation.

\textbf{Intraoral scans. }
We use two intraoral scan datasets with annotated faces: 3DTeethSeg \cite{ben20233dteethseg}, comprising 1{,}800 scans from 900 patients covering upper and lower jaws separately, and 3D-IOSSeg \cite{li2024fine}, comprising 180 scans with fine-grained teeth annotations. A key distinction is that 3DTeethSeg (and most prior open-source datasets) was acquired by \emph{indirect} scanning via a plaster base, whereas 3D-IOSSeg uses \emph{direct} intraoral scanning, yielding richer surrounding oral tissue and a distribution closer to the real clinical environment. Consequently, 3D-IOSSeg contains substantially more abnormal-tooth cases, raising additional segmentation challenges (see Figure~\ref{fig:datasets} in the appendix). As in previous work \cite{li2024fine,xi20253d}, we downsample the mesh to 16{,}000 faces.

\textbf{Intracranial aneurysm segments. }
We use the IntrA dataset \cite{yang2020intra} which contains 103 intracranial aneurysm surface meshes. Each mesh is derived from reconstructing scanned 2D magnetic resonance angiography (MRA) images, and its vertices are annotated with a binary segmentation of the aneurysm sac.
Full preprocessing details, including mesh cleanup, coordinate normalisation, spectral caching, and dataset-specific annotation conversions, are provided in Appendix~\ref{sec:app-data}.

\subsection{Mesh featurization}
We construct a three-tier per-vertex and per-edge feature set, combining intrinsic shape descriptors, local mesh structure, and dataset-specific global coordinate priors.
A central design requirement is that features should respect the symmetry group appropriate to each task.
Intracranial aneurysm meshes arrive in arbitrary orientation, so we target invariance to the full E(3) group (rotations, reflections, and translations).
Liver and dental meshes have clinically meaningful chirality: left versus right anatomy, so we target SE(3) invariance (rotations and translations only, no reflections). Full feature dimensions and per-dataset configurations are listed in Appendix~\ref{sec:app-features}.

\textbf{Intrinsic node features. }
The primary shape descriptor is the Heat Kernel Signature (HKS)~\citep{sun2009concise}, computed from the Laplace--Beltrami spectrum and sampled at 8 logarithmically spaced time scales, yielding an 8-dimensional per-vertex feature.
As an intrinsic quantity, HKS is invariant to the full E(3) group.
We additionally include pointwise area, the mean area of incident triangles, as a single-dimensional local surface-scale cue, which is likewise E(3)-invariant.

\textbf{Edge features. }
Each directed edge is assigned a dihedral angle between the normals of its two adjacent face pairs, encoding local surface curvature.
Dihedral angles are E(3)-invariant scalars.
Degree-normalised and precomputed edge weights (dataset-specific) are also included to modulate message aggregation.

\textbf{Global coordinate priors. }
Intrinsic descriptors capture local geometry but not global position within the organ, information that can be essential for tasks with strong spatial label priors.
For the dental datasets (3DTeethSeg and 3D-IOSSeg), we compute a PCA-derived anatomical frame from the dental arch and express vertex positions in cylindrical form $(r, \theta, z)$ (see Algorithm~\ref{alg:dental-frame} and the equivariance argument in Appendix~\ref{sec:app-features}).
The coordinates are invariant under SE(3) when the shape-derived frame rotates consistently with the mesh; the conditions for selecting such a frame are made explicit below.
For liver segmentation, we do the same construction, but use a different sign resolution strategy, and area-weighted quantities to alleviate tesselation sensitivity; see Appendix~\ref{sec:app-features} for details.

\textbf{PCA frame selection and rotation robustness. }
\label{sec:frame-selection}
PCA defines a set of candidate frames that transforms with the mesh, while obtaining invariant coordinates requires selecting a consistent member of that set.
For centred coordinates, a rotation $\mR$ transforms the covariance as $\mC \mapsto \mR\mC\mR^\top$.
With distinct, consistently ordered eigenvalues, the eigenvectors transform as $\mU \mapsto \mR\mU\mD$, where $\mD$ is a diagonal matrix of signs (this being the source of non-uniqueness).
Our dental frame resolves the anterior--posterior ($Y$) direction by arch concavity and enforces right-handedness, leaving the simultaneous $(X,Z)\mapsto(-X,-Z)$ choice.
The liver frame resolves two signs using area-weighted asymmetry and deterministic fallbacks, then derives the third axis by a cross product.
These frame-selection conditions, together with floating-point effects, explain why we report canonical and rotated dental/liver scores separately rather than asserting exact equality.
The fixed-angle evaluations show only small changes in overlap scores.
Appendix~\ref{sec:pca-sensitivity} measures changes in the PCA coordinates used as input features and in the resulting predictions after random-axis rotations.
Resolving the remaining dental sign choice is a principled refinement of the anatomical frame.

\subsection{Equivariant mesh encoder}

The EMNN architecture \cite{trang20243} extends EGNN \cite{satorras2021n} to triangle meshes by exploiting two kinds of geometric quantities: \emph{invariant} scalars---squared edge lengths $\|\vx_i - \vx_j\|^2$ and face areas $\|(\vx_j-\vx_i)\times(\vx_k-\vx_i)\|$---that serve as safe inputs to scalar feature updates, and \emph{equivariant} vectors---edge displacements $(\vx_i - \vx_j)$ and face normals $(\vx_j-\vx_i)\times(\vx_k-\vx_i)$---that appear exclusively in the coordinate update.
The edge-message update retains the EGNN form:
\begin{equation}
    \label{eq:mij-egnn}
    \vm_{ij} = \phi_e(\vh_i, \vh_j, \ve_{ij}, \|\vx_i - \vx_j\|^2).
\end{equation}
A surface-aware message from face $(i, j, k)$ to vertex $i$ is additionally defined as
\begin{equation}
    \label{eq:mijk-emnn}
    \vm_{ijk} = \phi_f(\vh_i, \vh_j + \vh_k, \|(\vx_j - \vx_i) \times (\vx_k - \vx_i)\|).
\end{equation}
The vertex feature update aggregates messages from neighbouring vertices and faces:
\begin{equation}
    \label{eq:hi-update}
    \vh_i = \phi_h\!\left(\vh_i,\; \sum_{j \in N(i)} \vm_{ij}, \sum_{(j, k) \in \tau(i)} \vm_{ijk}\right).
\end{equation}

The coordinate update couples the equivariant edge displacements and face normals:
\begin{equation}
    \label{eq:xi-update}
    \vx_i \leftarrow \vx_i + \sum_{j \in N(i)} (\vx_i - \vx_j)\, \phi_x(\vm_{ij}) + \sum_{(j, k) \in \tau(i)} \bigl((\vx_j - \vx_i) \times (\vx_k - \vx_i)\bigr)\, \phi_n(\vm_{ijk}).
\end{equation}
The first term propagates pairwise displacement information, while the second injects oriented local surface geometry through face normals without breaking equivariance.\footnote{For the face-normal branch, our E(3) claim follows the oriented-surface convention used in EMNN: each triangular face is ordered so that $(\vx_j-\vx_i) \times (\vx_k-\vx_i)$ points outward. Under a reflection, the embedded geometry changes handedness, so the reflected face must be rewound by swapping $j$ and $k$ to preserve the outward normal. This winding reversal cancels the sign flip of the cross product under reflections, yielding E(3)-equivariance for the oriented-mesh representation.}
All EAMS-family encoders add learned graph-level aggregation (i.e. a global node).

\subsection{Augmented message passing}
EMNN's local message passing limits the receptive field of a shallow encoder.
Beyond its shared global node, we study two multi-token mechanisms for longer-range context without deepening the network.

\paragraph{Soft regional aggregators.}
The soft regional aggregators (SRAs) partition the mesh into $K$ learnable region prototypes using a differentiable soft assignment.
At each forward pass, each vertex $i$ computes an assignment vector $\va_i = \mathrm{softmax}(\phi_a(\vh_i)) \in \mathbb{R}^K$; these stack row-wise into the \emph{assignment matrix} $\mA \in \mathbb{R}^{N \times K}$ with $[\mA]_{ik} = [\va_i]_k$.
The $k$-th region token is formed by weighted pooling: $\vr_k = \sum_{i} [\mA]_{ik}\, \vh_i$.
The region tokens are then mixed by a small transformer encoder before being scattered back to nodes (the full configuration is given in Appendix~\ref{sec:app-features}):
\(
    \vh_i \leftarrow \vh_i + \alpha \sum_{k} [\mA]_{ik}\, \phi_{\mathrm{proj}}(\hat{\vr}_k),
\)
where $\hat{\vr}_k$ denotes the transformer output for region $k$ and $\alpha$ is a learned residual weight.
This soft partition creates a structured compression bottleneck; it does not require spatially contiguous regions.
The full update procedure is given in Algorithm~\ref{alg:sr-update} (Appendix~\ref{sec:app-features}).

\paragraph{Virtual nodes.}
Virtual nodes~\citep[VN]{zhang2026fast} instead use an unpartitioned memory bank: $V$ virtual-node feature vectors $\{\vv_k\}$ are initialised from a shared learned seed and $V$ virtual-node coordinates $\{\vu_k\}$ are initialised at the graph centroid.
At each EMNN layer, real nodes exchange messages with virtual nodes as if they were additional graph neighbours, and virtual nodes aggregate information from all real nodes.
Unlike SRAs, virtual nodes impose no regional-partition assumption.
Algorithm~\ref{alg:fast-vn} (Appendix~\ref{sec:app-features}) gives the full layer update, showing the virtual-node exchange in the context of the complete EMNN block with edge and face-area branches; relative to the formulation in~\citep{zhang2026fast}, our model additionally includes the mesh-specific face-area coordinate update.

\subsection{Regularization objectives}
The augmented modules introduce additional latent structure, so we regularise them to keep their global summaries informative and stable during training.

\textbf{Boundary aware losses. }
Our supervised objective combines a prediction loss with a boundary-aware embedding loss. Specifically, we use the boundary difference over union \cite[bDoU]{sun2023boundary} loss for the segmentation predictions and add a contrastive boundary objective \cite[CBL]{tang2022contrastive} on the latent embeddings, encouraging nearby points across class boundaries to become more discriminative. We write this shared task loss as $\mathcal{L}_{\text{task}} = \mathcal{L}_{\text{pred}} + \lambda_{\text{cbl}}\,\mathcal{L}_{\text{cbl}}$.

\textbf{Regional assignment losses. }
For SRAs, we regularise the assignment matrix $\mA \in \mathbb{R}^{N \times K}$ (whose rows are the per-vertex assignment vectors $\va_i$) with two unsupervised objectives.
A \emph{diversity loss} penalises off-diagonal entries of the Gram matrix of the column-normalised assignments: $\mathcal{L}_{\text{div}} = \|\tilde{\mA}^\top \tilde{\mA} - \mI\|_F^2$, where $\tilde{\mA}$ denotes column-wise $\ell_2$-normalised $\mA$.
An \emph{equipartition loss} penalises imbalance in the total mass assigned to each region: $\mathcal{L}_{\text{eq}} = \mathrm{Var}(\vone^\top \mA) / (\bar{m}^2 + \eps)$, where $\bar{m}$ is the mean column mass.
The combined loss $\mathcal{L}_{\text{SRA}} = \lambda_{\text{div}}\,\mathcal{L}_{\text{div}} + \lambda_{\text{eq}}\,\mathcal{L}_{\text{eq}}$ is averaged per mesh.

\textbf{Virtual node losses. }
For virtual nodes, we use a kernel-based energy objective on the virtual-node coordinates $\{\vu_k\}$, following~\cite{zhang2026fast}. Let $k(\vp, \vq) = \exp(-\|\vp - \vq\|^2 / 2\sigma^2)$ be a Gaussian kernel, which is E(3)-invariant. We compute a virtual--virtual repulsion term $k_{\mathit{vv}}$ as the mean off-diagonal pairwise kernel among $\{\vu_k\}$, and a real--virtual attraction term $k_{\mathit{rv}}$ as the mean kernel between a random subsample of $\{\vx_i\}$ and $\{\vu_k\}$. The loss $\mathcal{L}_{\text{VN}} = w_{\mathit{vv}}\,k_{\mathit{vv}} - w_{\mathit{rv}}\,k_{\mathit{rv}}$ simultaneously discourages virtual nodes from collapsing together and encourages them to remain close to the mesh surface.

\textbf{Liver local continuity loss. }
Liver landmarks have subtle local geometric cues. We encourage spatially coherent predictions with a local continuity loss on adjacent edges. Let $\mathcal{A}$ denote the set of adjacent edge pairs. The loss $\mathcal{L}_{\text{cont}} = \frac{1}{|\mathcal{A}|} \sum_{(e_i, e_j) \in \mathcal{A}} \|\hat{y}_{e_i} - \hat{y}_{e_j}\|^2$ encourages smoothness in the predicted labels across neighboring edges.

\textbf{Final training objective. }
The full objective is obtained by adding the task-specific auxiliary terms to $\mathcal{L}_{\text{task}}$. For the base EAMS model, we optimise $\mathcal{L}_{\text{task}}$, and on liver meshes we additionally add the continuity term $\lambda_{\text{cont}}\,\mathcal{L}_{\text{cont}}$. For SRA+EAMS, we optimise
    $\mathcal{L}_{\text{task}} + \mathcal{L}_{\text{SRA}}$,
while for VN+EAMS we optimise
    $\mathcal{L}_{\text{task}} + \mathcal{L}_{\text{VN}}$.
All loss weights are given in Appendix~\ref{sec:app-training}.

\section{Experiments}
\label{sec:experiments}
All methods in the main comparison tables use the same bounded training rotations, with angles sampled uniformly from $[-0.15\pi,0.15\pi]$, identical dataset splits, and the same model-selection protocol.
Non-equivariant baselines are retrained with this augmentation; Fast-TGCN already uses it in its native pipeline.
Meshes are centred deterministically, so no separate translation augmentation is needed.
We evaluate canonical inputs and fixed 15\textdegree{} and 40\textdegree{} $z$-axis rotations; reflections are evaluated only on IntrA as a coordinate-handedness test, with face winding preserved under the oriented-surface convention.
Appendix~\ref{sec:random-axis} broadens this evaluation to random-axis rotations, while the qualitative figures use fixed-angle rotations for direct visual comparison.
Training and metric details are given in Appendix~\ref{sec:app-training}.

\begin{table*}[t]
\caption{%
  Intracranial aneurysm segmentation on the IntrA dataset \citep{yang2020intra}.
  Results are mean $\pm$ std over five cross-validation folds.
  \textbf{Bold}: best per column (all rows); \underline{underline}: second-best.
  EAMS, SRA+EAMS, and VN+EAMS are E(3)-invariant; their perturbation scores equal their baseline values and are marked $\dagger$.
  Full IoU under perturbation is in Table~\ref{tab:intra-robust-full}.
}
\label{tab:intra-main}
\tablestyle{4pt}{1.15}
\adjustbox{max width=\textwidth}{%
\begin{tabular}{lcccccccccc}
\toprule
& \multicolumn{5}{c}{Parent vessel} & \multicolumn{5}{c}{Aneurysm} \\
\cmidrule(lr){2-6}\cmidrule(lr){7-11}
& \multicolumn{2}{c}{Baseline} & \multicolumn{3}{c}{Perturbation Dice (\%) $\uparrow$} & \multicolumn{2}{c}{Baseline} & \multicolumn{3}{c}{Perturbation Dice (\%) $\uparrow$} \\
\cmidrule(lr){2-3}\cmidrule(lr){4-6}\cmidrule(lr){7-8}\cmidrule(lr){9-11}
Method & Dice (\%) $\uparrow$ & IoU (\%) $\uparrow$ & Rot-$z$ 15\textdegree{} & Rot-$z$ 40\textdegree{} & Refl-$x$ & Dice (\%) $\uparrow$ & IoU (\%) $\uparrow$ & Rot-$z$ 15\textdegree{} & Rot-$z$ 40\textdegree{} & Refl-$x$ \\
\midrule
DGCNN \citep{wang2019dynamic}
  & \pmval{97.16}{0.59} & \pmval{94.61}{1.06} & \pmval{96.91}{0.64} & \pmval{95.65}{0.19} & \pmval{95.85}{0.34}
  & \pmval{88.17}{1.69} & \pmval{80.59}{2.58} & \pmval{87.45}{1.71} & \pmval{81.75}{0.40} & \pmval{81.28}{2.76} \\
PTv3 \citep{wu2024point}
  & \pmval{95.28}{1.38} & \pmval{91.26}{2.41} & \pmval{95.11}{1.25} & \pmval{94.31}{1.05} & \pmval{93.67}{0.87}
  & \pmval{81.29}{5.28} & \pmval{72.30}{6.40} & \pmval{80.36}{4.96} & \pmval{77.24}{4.26} & \pmval{74.70}{4.44} \\
\midrule
\rowcolor{gray!10} EAMS
  & \pmval{98.31}{1.38} & \pmval{96.87}{2.59} & \pmval{98.31}{1.38}$^\dagger$ & \pmval{98.31}{1.38}$^\dagger$ & \pmval{98.31}{1.38}$^\dagger$
  & \pmval{93.59}{4.71} & \pmval{90.33}{7.72} & \pmval{93.59}{4.71}$^\dagger$ & \pmval{93.59}{4.71}$^\dagger$ & \pmval{93.59}{4.71}$^\dagger$ \\
\rowcolor{gray!10} SRA+EAMS
  & \underline{\pmval{98.69}{0.88}} & \underline{\pmval{97.71}{1.68}} & \underline{\pmval{98.69}{0.88}}$^\dagger$ & \underline{\pmval{98.69}{0.88}}$^\dagger$ & \underline{\pmval{98.69}{0.88}}$^\dagger$
  & \textbf{\pmval{95.81}{2.90}} & \textbf{\pmval{93.32}{5.18}} & \textbf{\pmval{95.81}{2.90}}$^\dagger$ & \textbf{\pmval{95.81}{2.90}}$^\dagger$ & \textbf{\pmval{95.81}{2.90}}$^\dagger$ \\
\rowcolor{gray!10} VN+EAMS
  & \textbf{\pmval{98.90}{0.98}} & \textbf{\pmval{97.95}{1.84}} & \textbf{\pmval{98.90}{0.98}}$^\dagger$ & \textbf{\pmval{98.90}{0.98}}$^\dagger$ & \textbf{\pmval{98.90}{0.98}}$^\dagger$
  & \underline{\pmval{95.31}{4.55}} & \underline{\pmval{93.15}{6.54}} & \underline{\pmval{95.31}{4.55}}$^\dagger$ & \underline{\pmval{95.31}{4.55}}$^\dagger$ & \underline{\pmval{95.31}{4.55}}$^\dagger$ \\
\bottomrule
\end{tabular}%
}
\end{table*}

\subsection{Intracranial aneurysm segmentation}

We evaluate on the IntrA dataset \citep{yang2020intra}, which comprises 103 intracranial aneurysm surface meshes with per-vertex binary labels separating the aneurysm sac from the parent vessel wall.
We compare against PointTransformerV3 (PTv3) \citep{wu2024point} and DGCNN \citep{wang2019dynamic} as baseline methods.
Both baselines operate natively on point clouds and are provided the same vertex coordinates as input; our EAMS variants operate directly on the mesh graph.
All results are averaged over five cross-validation folds.

Table~\ref{tab:intra-main} reports Dice and IoU for both segmentation targets on unperturbed meshes alongside Dice under test-time geometric perturbations.
All three EAMS variants improve consistently over both matched-augmentation baselines across every metric, especially under rotation: at 40\textdegree{}, aneurysm Dice is 93.59--95.81\% for EAMS variants versus 81.75\% for the strongest baseline.
Full IoU results are provided in Appendix Table~\ref{tab:intra-robust-full}; Figure~\ref{fig:intra-qual} compares separate sets of representative canonical and 40\textdegree{}-rotated cases.

\begin{figure*}[tbp]
\centering
\makebox[0.245\textwidth][c]{\textbf{Ground truth}}\hfill
\makebox[0.245\textwidth][c]{\textbf{DGCNN}}\hfill
\makebox[0.245\textwidth][c]{\textbf{PTv3}}\hfill
\makebox[0.245\textwidth][c]{\textbf{VN+EAMS}}

\par\smallskip

\begin{subfigure}[t]{0.245\textwidth}
  \qualimgintra{intra/emnn/base/AN54-1_full_edge_labels_xz_gt}
\end{subfigure}\hfill
\begin{subfigure}[t]{0.245\textwidth}
  \qualimgintra{intra/dgcnn/base/AN54-1_full_edge_labels_xz_pred}
\end{subfigure}\hfill
\begin{subfigure}[t]{0.245\textwidth}
  \qualimgintra{intra/ptv3/base/AN54-1_full_edge_labels_xz_pred}
\end{subfigure}\hfill
\begin{subfigure}[t]{0.245\textwidth}
  \qualimgintra{intra/vn_emnn/base/AN54-1_full_edge_labels_xz_pred}
\end{subfigure}

\par\smallskip
\begin{subfigure}[t]{0.245\textwidth}
  \qualimgintra{intra/emnn/base/AN193-1_full_edge_labels_xz_gt}
\end{subfigure}\hfill
\begin{subfigure}[t]{0.245\textwidth}
  \qualimgintra{intra/dgcnn/base/AN193-1_full_edge_labels_xz_pred}
\end{subfigure}\hfill
\begin{subfigure}[t]{0.245\textwidth}
  \qualimgintra{intra/ptv3/base/AN193-1_full_edge_labels_xz_pred}
\end{subfigure}\hfill
\begin{subfigure}[t]{0.245\textwidth}
  \qualimgintra{intra/vn_emnn/base/AN193-1_full_edge_labels_xz_pred}
\end{subfigure}

\par\smallskip
\begin{subfigure}[t]{0.245\textwidth}
  \qualimgintra{intra/emnn/base/AN195_full_edge_labels_xz_gt}
\end{subfigure}\hfill
\begin{subfigure}[t]{0.245\textwidth}
  \qualimgintra{intra/dgcnn/base/AN195_full_edge_labels_xz_pred}
\end{subfigure}\hfill
\begin{subfigure}[t]{0.245\textwidth}
  \qualimgintra{intra/ptv3/base/AN195_full_edge_labels_xz_pred}
\end{subfigure}\hfill
\begin{subfigure}[t]{0.245\textwidth}
  \qualimgintra{intra/vn_emnn/base/AN195_full_edge_labels_xz_pred}
\end{subfigure}


\par\medskip\hrule\par\smallskip

\begin{subfigure}[t]{0.245\textwidth}
  \qualimgintra{intra/emnn/rotate_40/AN125_full_edge_labels_xz_gt}
\end{subfigure}\hfill
\begin{subfigure}[t]{0.245\textwidth}
  \qualimgintra{intra/dgcnn/rotate_40/AN125_full_edge_labels_xz_pred}
\end{subfigure}\hfill
\begin{subfigure}[t]{0.245\textwidth}
  \qualimgintra{intra/ptv3/rotate_40/AN125_full_edge_labels_xz_pred}
\end{subfigure}\hfill
\begin{subfigure}[t]{0.245\textwidth}
  \qualimgintra{intra/vn_emnn/rotate_40/AN125_full_edge_labels_xz_pred}
\end{subfigure}

\par\smallskip
\begin{subfigure}[t]{0.245\textwidth}
  \qualimgintra{intra/emnn/rotate_40/AN178_full_edge_labels_xz_gt}
\end{subfigure}\hfill
\begin{subfigure}[t]{0.245\textwidth}
  \qualimgintra{intra/dgcnn/rotate_40/AN178_full_edge_labels_xz_pred}
\end{subfigure}\hfill
\begin{subfigure}[t]{0.245\textwidth}
  \qualimgintra{intra/ptv3/rotate_40/AN178_full_edge_labels_xz_pred}
\end{subfigure}\hfill
\begin{subfigure}[t]{0.245\textwidth}
  \qualimgintra{intra/vn_emnn/rotate_40/AN178_full_edge_labels_xz_pred}
\end{subfigure}

\par\smallskip
\begin{subfigure}[t]{0.245\textwidth}
  \qualimgintra{intra/emnn/rotate_40/AN196-2_full_edge_labels_xz_gt}
\end{subfigure}\hfill
\begin{subfigure}[t]{0.245\textwidth}
  \qualimgintra{intra/dgcnn/rotate_40/AN196-2_full_edge_labels_xz_pred}
\end{subfigure}\hfill
\begin{subfigure}[t]{0.245\textwidth}
  \qualimgintra{intra/ptv3/rotate_40/AN196-2_full_edge_labels_xz_pred}
\end{subfigure}\hfill
\begin{subfigure}[t]{0.245\textwidth}
  \qualimgintra{intra/vn_emnn/rotate_40/AN196-2_full_edge_labels_xz_pred}
\end{subfigure}

\caption{IntrA comparisons on representative meshes, with top half in the canonical pose and bottom half after a 40\textdegree{} $z$-axis rotation. Appendix Figure~\ref{fig:intra-qual-ours} adds EAMS and SRA+EAMS predictions for these cases.}
\label{fig:intra-qual}
\end{figure*}


\subsection{Intraoral scan tooth segmentation}

We evaluate on two intraoral scan benchmarks sharing the same FDI labelling scheme.
\textbf{3D-IOSSeg} \citep{li2024fine} comprises 180 direct intraoral scans with per-face labels spanning the gingiva and 16 tooth classes, with results averaged over five cross-validation folds.
\textbf{Teeth3DS} \citep{ben20233dteethseg} is a larger benchmark of 1800 upper and lower jaw scans from the 3DTeethSeg'22 challenge; owing to its size we train on a single fold on its designated split of 1200 training and 600 test scans.
Table~\ref{tab:iosseg-main} reports Average IoU for both datasets: we average IoU over the classes present in each mesh and then average those per-mesh values over the evaluation set, rather than averaging the final per-class IoU summaries. Following common practice we squash mirrored FDI pair labels (e.g.\ T11/31).
We compare against DGCNN \citep{wang2019dynamic}, PTv3 \citep{wu2024point}, and Fast-TGCN \citep{li2024fine}; on Teeth3DS we retain Fast-TGCN as the strongest baseline.
Matched-augmentation per-class results for 3D-IOSSeg are provided in Tables~\ref{tab:iosseg-pertooth-left} and~\ref{tab:iosseg-pertooth-right}; Tables~\ref{tab:teeth3ds-pertooth-left} and~\ref{tab:teeth3ds-pertooth-right} give the Teeth3DS breakdown.

On 3D-IOSSeg and Teeth3DS, VN+EAMS approaches the best baseline Fast-TGCN despite being nearly $\mathbf{10\times}$ \textbf{smaller in parameter count} (1.9M vs. 19.8M parameters).
Performance on the rearmost wisdom-tooth classes (T18/38 and T28/48) is the main driver of variance across methods, owing to greater morphological variation and fewer training examples. The robustness gap persists under matched augmentation: on 3D-IOSSeg, the strongest baseline falls from 81.05\% to 52.33\% IoU at 40\textdegree{}, while VN+EAMS changes from 79.74\% to 80.09\%. The small variation in EAMS-family scores is interpreted through the frame-selection conditions in Section~\ref{sec:frame-selection}, with a dedicated sensitivity analysis in Appendix~\ref{sec:pca-sensitivity}. Figure~\ref{fig:teeth-qual} shows the same trend qualitatively: Fast-TGCN is competitive at baseline but deteriorates under rotation, whereas VN+EAMS remains stable. 

\begin{table*}[t]
\caption{%
  Intraoral scan tooth segmentation on 3D-IOSSeg \citep{li2024fine} and Teeth3DS \citep{ben20233dteethseg}.
  Metric is Average IoU (\%), obtained by averaging each mesh over its ground-truth-present classes and then averaging over meshes, not by averaging the final per-class IoU summaries.
  3D-IOSSeg results are mean $\pm$ std over five cross-validation folds; Teeth3DS results are from a single fold (no std).
  \textbf{Bold}: best per column (all rows); \underline{underline}: second-best.
  The frame-selection conditions and sources of variation between canonical and rotated scores are explained in Section~\ref{sec:frame-selection}; we report each perturbation empirically.
  Per-class results are in Tables~\ref{tab:iosseg-pertooth-left}~and~\ref{tab:iosseg-pertooth-right}.
}
\label{tab:iosseg-main}
\tablestyle{6pt}{1.15}
\adjustbox{max width=\textwidth}{%
\begin{tabular}{lcccccc}
\toprule
& \multicolumn{3}{c}{3D-IOSSeg \citep{li2024fine}} & \multicolumn{3}{c}{Teeth3DS \citep{ben20233dteethseg}} \\
\cmidrule(lr){2-4}\cmidrule(lr){5-7}
Method & Baseline $\uparrow$ & Rot-$z$ 15\textdegree{} $\uparrow$ & Rot-$z$ 40\textdegree{} $\uparrow$
      & Baseline $\uparrow$ & Rot-$z$ 15\textdegree{} $\uparrow$ & Rot-$z$ 40\textdegree{} $\uparrow$ \\
\midrule
DGCNN \citep{wang2019dynamic} & \pmval{65.92}{0.92} & \pmval{58.08}{0.72} & \pmval{37.92}{0.49} & --- & --- & --- \\
PTv3 \citep{wu2024point}      & \underline{\pmval{80.45}{0.40}} & \underline{\pmval{77.61}{0.32}} & \pmval{50.16}{2.39} & --- & --- & --- \\
Fast-TGCN \citep{li2024fine}  & \textbf{\pmval{81.05}{0.74}} & \pmval{76.45}{1.14} & \pmval{52.33}{1.62} & \textbf{84.31} & 81.84 & 58.53 \\
\midrule
\rowcolor{gray!10} EAMS      & \pmval{71.43}{1.64} & \pmval{71.71}{1.53} & \pmval{72.05}{1.60} & 80.54 & 80.94 & 81.11 \\
\rowcolor{gray!10} SRA+EAMS  & \pmval{75.28}{1.21} & \pmval{75.49}{1.44} & \underline{\pmval{75.98}{1.23}} & 82.31 & \underline{82.98} & \underline{83.05} \\
\rowcolor{gray!10} VN+EAMS   & \pmval{79.74}{0.86} & \textbf{\pmval{79.92}{0.98}} & \textbf{\pmval{80.09}{1.01}} & \underline{83.16} & \textbf{83.33} & \textbf{83.40} \\
\bottomrule
\end{tabular}%
}
\end{table*}

\begin{figure*}[!t]
\centering
\makebox[0.153\textwidth][c]{\textbf{GT}}\hfill
\makebox[0.153\textwidth][c]{\textbf{Fast-TGCN}}\hfill
\makebox[0.153\textwidth][c]{\textbf{VN+EAMS}}\hspace{0.004\textwidth}\vrule width 0.6pt\hspace{0.004\textwidth}
\makebox[0.153\textwidth][c]{\textbf{GT}}\hfill
\makebox[0.153\textwidth][c]{\textbf{Fast-TGCN}}\hfill
\makebox[0.153\textwidth][c]{\textbf{VN+EAMS}}

\par\vspace{2pt}

\begin{subfigure}[t]{0.153\textwidth}
  \qualimgteeth{teeth/vn_emnn/base/003_U_face_labels_xz_gt}
\end{subfigure}\hfill
\begin{subfigure}[t]{0.153\textwidth}
  \qualimgteeth{teeth/tcgn/base/003_U_face_labels_xz_pred}
\end{subfigure}\hfill
\begin{subfigure}[t]{0.153\textwidth}
  \qualimgteeth{teeth/vn_emnn/base/003_U_face_labels_xz_pred}
\end{subfigure}\hspace{0.004\textwidth}\vrule width 0.6pt\hspace{0.004\textwidth}
\begin{subfigure}[t]{0.153\textwidth}
  \qualimgteeth{teeth/vn_emnn/rotate_40/003_U_face_labels_xz_gt}
\end{subfigure}\hfill
\begin{subfigure}[t]{0.153\textwidth}
  \qualimgteeth{teeth/tcgn/rotate_40/003_U_face_labels_xz_pred}
\end{subfigure}\hfill
\begin{subfigure}[t]{0.153\textwidth}
  \qualimgteeth{teeth/vn_emnn/rotate_40/003_U_face_labels_xz_pred}
\end{subfigure}

\par\vspace{1pt}
\begin{subfigure}[t]{0.153\textwidth}
  \qualimgteeth{teeth/vn_emnn/base/006_L_face_labels_xz_gt}
\end{subfigure}\hfill
\begin{subfigure}[t]{0.153\textwidth}
  \qualimgteeth{teeth/tcgn/base/006_L_face_labels_xz_pred}
\end{subfigure}\hfill
\begin{subfigure}[t]{0.153\textwidth}
  \qualimgteeth{teeth/vn_emnn/base/006_L_face_labels_xz_pred}
\end{subfigure}\hspace{0.004\textwidth}\vrule width 0.6pt\hspace{0.004\textwidth}
\begin{subfigure}[t]{0.153\textwidth}
  \qualimgteeth{teeth/vn_emnn/rotate_40/006_L_face_labels_xz_gt}
\end{subfigure}\hfill
\begin{subfigure}[t]{0.153\textwidth}
  \qualimgteeth{teeth/tcgn/rotate_40/006_L_face_labels_xz_pred}
\end{subfigure}\hfill
\begin{subfigure}[t]{0.153\textwidth}
  \qualimgteeth{teeth/vn_emnn/rotate_40/006_L_face_labels_xz_pred}
\end{subfigure}

\par\vspace{1pt}
\begin{subfigure}[t]{0.153\textwidth}
  \qualimgteeth{teeth/vn_emnn/base/006_U_face_labels_xz_gt}
\end{subfigure}\hfill
\begin{subfigure}[t]{0.153\textwidth}
  \qualimgteeth{teeth/tcgn/base/006_U_face_labels_xz_pred}
\end{subfigure}\hfill
\begin{subfigure}[t]{0.153\textwidth}
  \qualimgteeth{teeth/vn_emnn/base/006_U_face_labels_xz_pred}
\end{subfigure}\hspace{0.004\textwidth}\vrule width 0.6pt\hspace{0.004\textwidth}
\begin{subfigure}[t]{0.153\textwidth}
  \qualimgteeth{teeth/vn_emnn/rotate_40/006_U_face_labels_xz_gt}
\end{subfigure}\hfill
\begin{subfigure}[t]{0.153\textwidth}
  \qualimgteeth{teeth/tcgn/rotate_40/006_U_face_labels_xz_pred}
\end{subfigure}\hfill
\begin{subfigure}[t]{0.153\textwidth}
  \qualimgteeth{teeth/vn_emnn/rotate_40/006_U_face_labels_xz_pred}
\end{subfigure}

\caption{%
  Qualitative tooth-segmentation comparisons, with the left half showing the canonical orientation and the right half showing a 40\textdegree{} $z$-axis rotation.
}
\label{fig:teeth-qual}
\end{figure*}

\FloatBarrier

\subsection{Liver surface segmentation}

We evaluate on the liver surface segmentation dataset of \citep{zhang2025nested}, which comprises surface meshes of the liver with per-edge labels spanning three anatomical classes: background, ligament, and ridge.
We compare against PointNet++ \citep{qi2017pointnet++}, DGCNN \citep{wang2019dynamic}, and MeshGraphCNN \citep{zhang2025nested} as baselines.
All three EAMS variants operate directly on the mesh graph and target SE(3)-invariance; for liver, SRA+EAMS uses $K=32$ regional tokens and VN+EAMS uses $V=8$ virtual nodes.
This task highlights a balanced trade-off between canonical-pose accuracy and rotation robustness: the baselines can exploit the near-canonical scan alignment in the training set as an orientation prior, whereas EAMS combines intrinsic geometry with shape-derived anatomical coordinates.

\begin{figure*}[!t]
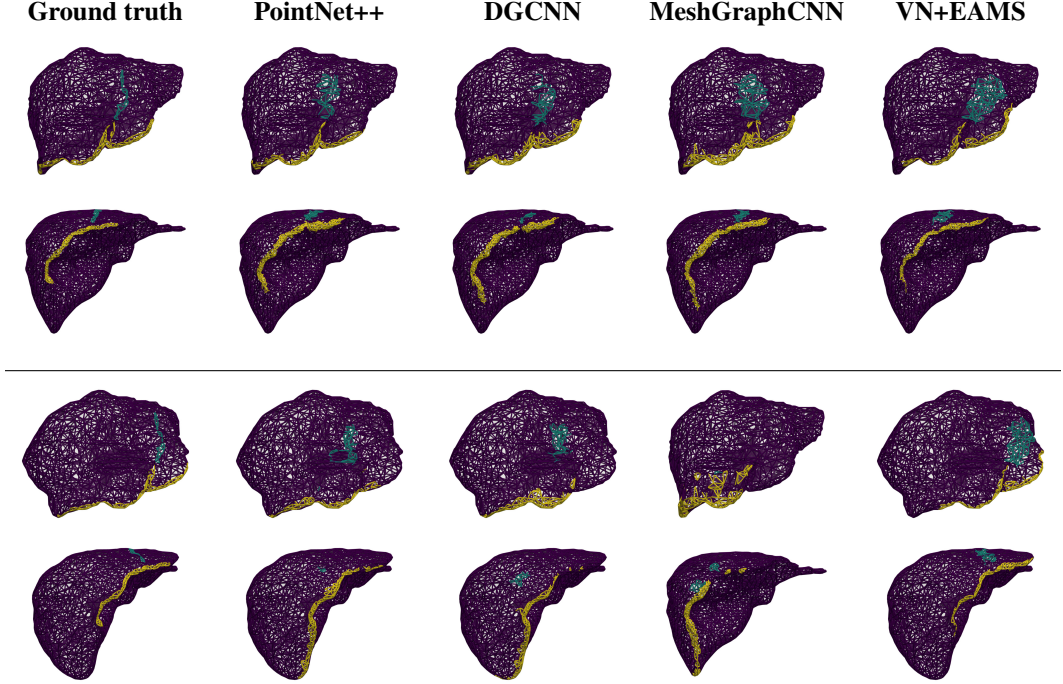

\centering
\makebox[0.19\textwidth][c]{\textbf{Ground truth}}\hfill
\makebox[0.19\textwidth][c]{\textbf{PointNet++}}\hfill
\makebox[0.19\textwidth][c]{\textbf{DGCNN}}\hfill
\makebox[0.19\textwidth][c]{\textbf{MeshGraphCNN}}\hfill
\makebox[0.19\textwidth][c]{\textbf{VN+EAMS}}

\par\smallskip

\begin{subfigure}[t]{0.19\textwidth}
  \qualimgliver{liver/pointnet2/base/3Dircadb-10_face_labels_xz_gt}
\end{subfigure}\hfill
\begin{subfigure}[t]{0.19\textwidth}
  \qualimgliver{liver/pointnet2/base/3Dircadb-10_face_labels_xz_pred}
\end{subfigure}\hfill
\begin{subfigure}[t]{0.19\textwidth}
  \qualimgliver{liver/dgcnn/base/3Dircadb-10_face_labels_xz_pred}
\end{subfigure}\hfill
\begin{subfigure}[t]{0.19\textwidth}
  \qualimgliver{liver/mesh_graph_cnn/base/3Dircadb-10_face_labels_xz_pred}
\end{subfigure}\hfill
\begin{subfigure}[t]{0.19\textwidth}
  \qualimgliver{liver/vn_emnn_final/base/3Dircadb-10_face_labels_xz_pred}
\end{subfigure}

\par\smallskip
\begin{subfigure}[t]{0.19\textwidth}
  \qualimgliver{liver/pointnet2/base/LiTS-10_face_labels_xz_gt}
\end{subfigure}\hfill
\begin{subfigure}[t]{0.19\textwidth}
  \qualimgliver{liver/pointnet2/base/LiTS-10_face_labels_xz_pred}
\end{subfigure}\hfill
\begin{subfigure}[t]{0.19\textwidth}
  \qualimgliver{liver/dgcnn/base/LiTS-10_face_labels_xz_pred}
\end{subfigure}\hfill
\begin{subfigure}[t]{0.19\textwidth}
  \qualimgliver{liver/mesh_graph_cnn/base/LiTS-10_face_labels_xz_pred}
\end{subfigure}\hfill
\begin{subfigure}[t]{0.19\textwidth}
  \qualimgliver{liver/vn_emnn_final/base/LiTS-10_face_labels_xz_pred}
\end{subfigure}

\par\smallskip
\par\medskip
\hrule
\par\smallskip

\begin{subfigure}[t]{0.19\textwidth}
  \qualimgliver{liver/pointnet2/rotate_40/3Dircadb-10_face_labels_xz_gt}
\end{subfigure}\hfill
\begin{subfigure}[t]{0.19\textwidth}
  \qualimgliver{liver/pointnet2/rotate_40/3Dircadb-10_face_labels_xz_pred}
\end{subfigure}\hfill
\begin{subfigure}[t]{0.19\textwidth}
  \qualimgliver{liver/dgcnn/rotate_40/3Dircadb-10_face_labels_xz_pred}
\end{subfigure}\hfill
\begin{subfigure}[t]{0.19\textwidth}
  \qualimgliver{liver/mesh_graph_cnn/rotate_40/3Dircadb-10_face_labels_xz_pred}
\end{subfigure}\hfill
\begin{subfigure}[t]{0.19\textwidth}
  \qualimgliver{liver/vn_emnn_final/rotate_40/3Dircadb-10_face_labels_xz_pred}
\end{subfigure}

\par\smallskip
\begin{subfigure}[t]{0.19\textwidth}
  \qualimgliver{liver/pointnet2/rotate_40/LiTS-10_face_labels_xz_gt}
\end{subfigure}\hfill
\begin{subfigure}[t]{0.19\textwidth}
  \qualimgliver{liver/pointnet2/rotate_40/LiTS-10_face_labels_xz_pred}
\end{subfigure}\hfill
\begin{subfigure}[t]{0.19\textwidth}
  \qualimgliver{liver/dgcnn/rotate_40/LiTS-10_face_labels_xz_pred}
\end{subfigure}\hfill
\begin{subfigure}[t]{0.19\textwidth}
  \qualimgliver{liver/mesh_graph_cnn/rotate_40/LiTS-10_face_labels_xz_pred}
\end{subfigure}\hfill
\begin{subfigure}[t]{0.19\textwidth}
  \qualimgliver{liver/vn_emnn_final/rotate_40/LiTS-10_face_labels_xz_pred}
\end{subfigure}

\par\smallskip
\caption{%
  Qualitative liver-surface comparisons on two representative meshes, with the top half in the canonical orientation and the bottom half after a 40\textdegree{} $z$-axis rotation. MeshGraphCNN uses a slightly different liver mesh because of its preprocessing requirements (Appendix~\ref{sec:app-data}). Appendix Figure~\ref{fig:liver-qual-ours} qualitatively compares three EAMS-family columns.
}
\label{fig:liver-qual}
\end{figure*}

Table~\ref{tab:liver-main} focuses on the main segmentation metric, Dice, together with Chamfer distance (CD) for the ligament and ridge classes on unperturbed meshes and under a 40\textdegree{} $z$-axis rotation. We report CD after multiplying by $100$ for readability. EAMS, SRA+EAMS, and VN+EAMS remain nearly unchanged under rotation, while the non-equivariant baselines degrade sharply. Full IoU, CD$\times 100$, and Hausdorff distance (HD) results for the baseline, 15\textdegree{}, and 40\textdegree{} conditions are given in Table~\ref{tab:liver-robust-full}, and Figure~\ref{fig:liver-qual} shows representative baseline and 40\textdegree{}-rotated cases.

\begin{table*}[t]
\caption{%
  Liver surface segmentation on the dataset of \citep{zhang2025nested}.
  Results are mean $\pm$ std over five cross-validation folds.
  CD is reported after multiplying by $100$.
  \textbf{Bold}: best per column (all rows); \underline{underline}: second-best.
  $\dagger$: MeshGraphCNN results are our own reproduction.
  Full IoU, CD$\times 100$, and HD results are in Table~\ref{tab:liver-robust-full}; frame-selection conditions are explained in Section~\ref{sec:frame-selection}.
}
\label{tab:liver-main}
\tablestyle{5pt}{1.15}
\adjustbox{max width=\textwidth}{%
\begin{tabular}{lcccccccc}
\toprule
& \multicolumn{4}{c}{Ligament} & \multicolumn{4}{c}{Ridge} \\
\cmidrule(lr){2-5}\cmidrule(lr){6-9}
& \multicolumn{2}{c}{Baseline} & \multicolumn{2}{c}{Rot-$z$ 40\textdegree{}} & \multicolumn{2}{c}{Baseline} & \multicolumn{2}{c}{Rot-$z$ 40\textdegree{}} \\
\cmidrule(lr){2-3}\cmidrule(lr){4-5}\cmidrule(lr){6-7}\cmidrule(lr){8-9}
Method & Dice (\%) $\uparrow$ & CD$\times 100$ $\downarrow$ & Dice (\%) $\uparrow$ & CD$\times 100$ $\downarrow$
      & Dice (\%) $\uparrow$ & CD$\times 100$ $\downarrow$ & Dice (\%) $\uparrow$ & CD$\times 100$ $\downarrow$ \\
\midrule
PointNet++ \citep{qi2017pointnet++}
  & \underline{\pmval{36.18}{2.54}} & \textbf{\pmval{0.220}{0.030}} & \pmval{11.79}{2.21} & \pmval{2.105}{1.187}
  & \underline{\pmval{59.19}{0.74}} & \textbf{\pmval{0.099}{0.016}} & \pmval{47.89}{1.26} & \pmval{0.879}{0.132} \\
DGCNN \citep{wang2019dynamic}
  & \pmval{33.12}{2.78} & \underline{\pmval{0.416}{0.028}} & \pmval{7.47}{0.89} & \pmval{3.237}{0.797}
  & \pmval{58.20}{2.60} & \underline{\pmval{0.105}{0.032}} & \pmval{42.42}{0.59} & \pmval{1.457}{0.169} \\
MeshGraphCNN$^\dagger$ \citep{zhang2025nested}
  & \textbf{\pmval{44.13}{0.94}} & \pmval{0.658}{0.030} & \pmval{6.21}{0.44} & \pmval{6.595}{0.135}
  & \textbf{\pmval{59.87}{0.24}} & \pmval{0.245}{0.009} & \pmval{40.08}{0.16} & \pmval{2.439}{0.021} \\
\midrule
\rowcolor{gray!10} EAMS
  & \pmval{20.46}{1.25} & \pmval{3.928}{0.668} & \pmval{20.47}{1.26} & \pmval{4.307}{0.193}
  & \pmval{56.36}{0.82} & \pmval{0.873}{0.064} & \underline{\pmval{56.35}{0.82}} & \underline{\pmval{0.876}{0.065}} \\
\rowcolor{gray!10} SRA+EAMS
  & \pmval{26.33}{2.16} & \pmval{1.722}{0.651} & \textbf{\pmval{26.32}{2.16}} & \textbf{\pmval{1.722}{0.650}}
  & \pmval{54.54}{1.11} & \pmval{1.071}{0.372} & \pmval{54.52}{1.14} & \pmval{1.069}{0.371} \\
\rowcolor{gray!10} VN+EAMS
  & \pmval{20.98}{1.38} & \pmval{2.013}{0.592} & \underline{\pmval{20.99}{1.38}} & \underline{\pmval{2.012}{0.591}}
  & \pmval{58.50}{2.07} & \pmval{0.610}{0.160} & \textbf{\pmval{58.51}{2.08}} & \textbf{\pmval{0.609}{0.159}} \\
\bottomrule
\end{tabular}%
}
\end{table*}

Accordingly, the EAMS family trails the strongest baselines on unperturbed meshes---particularly for ligament Dice---but remains nearly unchanged under rotation, unlike all three baselines. The ligament and ridge occupy anatomically consistent frontal regions of the liver, so orientation-aware models can use pose as a strong cue, whereas EAMS uses anatomy-relative position rather than raw coordinate pose. The raw-coordinate diagnostic in Appendix~\ref{sec:pose-diagnostic} supports the utility of consistent dataset alignment; the PCA-coordinate analysis in Appendix~\ref{sec:pca-sensitivity} shows stable liver coordinates and predictions under the evaluated rotations. Figure~\ref{fig:liver-qual} shows competitive qualitative structure. Among the EAMS variants, SRA+EAMS gives the strongest rotated ligament Dice and CD, while VN+EAMS gives the best rotated ridge Dice and CD; Table~\ref{tab:liver-robust-full} shows the same stability pattern across the full baseline, 15\textdegree{}, and 40\textdegree{} evaluation suite.

\subsection{Ablation studies}

We build on the base encoder step by step to examine which additions improve segmentation.
Tables~\ref{tab:context-ablations} and~\ref{tab:liver-ablations} report single-fold additive ablations for vertex-level IntrA, face-level 3D-IOSSeg, and edge-level liver supervision, respectively.

\paragraph{Embedding supervision and anatomical context.}
Adding contrastive boundary loss (CBL) on embeddings improves both IntrA and 3D-IOSSeg while retaining the same prediction loss.
The task-specific EAMS base then adds the global node on IntrA and the global node together with PCA coordinates on 3D-IOSSeg; this combined dental intervention raises IoU from 32.50\% to 71.64\%.
The liver ablation follows the same additive logic: augmenting intrinsic features with the anatomical frame raises ligament Dice from 12.76\% to 18.78\% and ridge Dice from 51.06\% to 56.17\%.
The local continuity loss further improves ligament segmentation, with little benefit to the ridge.
Together, these comparisons show the value of combining local shape cues with anatomical context across supervision types.

\paragraph{Global message passing and regularization.}
Starting from the task-specific EAMS bases, both SRA and VN improve IntrA and 3D-IOSSeg even without their auxiliary losses.
The full VN objective improves both tasks further, while SRA regularization improves 3D-IOSSeg but lowers IntrA scores relative to unregularized SRA.
A width-160 parameter-matched EMNN (\emph{Wide EMNN} in Table~\ref{tab:context-ablations}) falls short of both full context variants: widening alone does not reproduce their gains in these single-fold comparisons.
Appendix~\ref{sec:feature-ablations} complements this additive analysis by removing intrinsic features and the global node, while Appendix~\ref{sec:token-ablations} examines token-count sensitivity.

\begin{table*}[t]
\centering
\begin{minipage}[t]{0.55\textwidth}
\captionof{table}{Single-fold IntrA (vertex) and IOSSeg (face) ablations. \emph{No reg.}: module auxiliary loss removed.}
\label{tab:context-ablations}
\tablestyle{2pt}{1.1}
\begin{tabular}{lccc}
\toprule
& \multicolumn{2}{c}{IntrA} & IOSSeg \\
\cmidrule(lr){2-3}
Variant & Dice (\%) $\uparrow$ & IoU (\%) $\uparrow$ & IoU (\%) $\uparrow$ \\
\midrule
EMNN w/o CBL & 91.03 & 85.22 & 29.97 \\
+ CBL & 91.56 & 85.81 & 32.50 \\
EAMS base & 92.36 & 86.92 & 71.64 \\
\midrule
+ SRA, no reg. & 93.79 & 89.33 & 74.49 \\
+ full SRA & 92.97 & 88.19 & 75.61 \\
+ VN, no reg. & 93.39 & 90.16 & 80.65 \\
\rowcolor{gray!10} + full VN & \textbf{94.51} & \textbf{91.28} & \textbf{81.74} \\
\midrule
Wide EMNN & 90.40 & 84.31 & 72.22 \\
\bottomrule
\end{tabular}
\end{minipage}\hfill
\begin{minipage}[t]{0.42\textwidth}
\captionof{table}{Single-fold liver (edge) ablations.}
\label{tab:liver-ablations}
\tablestyle{2pt}{1.1}
\begin{tabular}{llcc}
\toprule
Class & Variant & Dice (\%) $\uparrow$ & CD $\times 100$ $\downarrow$ \\
\midrule
Ligament & Base & 12.76 & 5.78 \\
 & + frame & 18.78 & 4.38 \\
 & + cont. & 19.51 & 3.81 \\
\midrule
Ridge & Base & 51.06 & 1.70 \\
 & + frame & 56.17 & 0.81 \\
 & + cont. & 56.08 & 0.88 \\
\bottomrule
\end{tabular}
\end{minipage}
\end{table*}

\section{Discussion}
\label{sec:discussion}
EAMS demonstrates that a lightweight shared mesh encoder can support anatomical segmentation across edge-, vertex-, and face-level supervision without relying on a separate backbone for each task.
Across four dataset settings, it combines intrinsic geometry, anatomy-relative coordinates, and global context while retaining fewer than 2M parameters.
Under matched training augmentation, EAMS variants improve both canonical and perturbed IntrA segmentation; VN+EAMS approaches the specialized dental baseline with 1.9M versus 19.8M parameters and remains stable under fixed-angle rotations (Tables~\ref{tab:intra-main} and~\ref{tab:iosseg-main}).
The advantage also persists in the broader random-axis tests (Appendix~\ref{sec:random-axis}), showing that it is not explained by unequal orientation augmentation.

The ablations identify complementary roles for local geometry, anatomical position, and global aggregation.
Removing intrinsic descriptors weakens segmentation, while adding anatomy-relative context improves the dental and liver bases (Section~\ref{sec:experiments}; Appendix~\ref{sec:feature-ablations}).
Both global-context mechanisms improve the common IntrA and IOSSeg bases, and widening EMNN alone does not reproduce those gains.
Full VN leads the single-fold component comparison, whereas SRA offers structured regional compression and leads five-fold aneurysm segmentation.
These findings provide design guidance: VN regularization improves both tested tasks, while SRA regularization is task-dependent and should be selected by validation.

A further contribution is the distinction between arbitrary coordinate pose and useful acquisition alignment.
The raw-XYZ diagnostic (Appendix~\ref{sec:pose-diagnostic}) shows that pose-sensitive inputs help liver and dental segmentation but hurt IntrA.
Together with the anatomy-relative feature ablations, this supports recovering useful organ-relative position rather than discarding global information when removing arbitrary pose, and motivates hybrid pose-aware/equivariant models where alignment is informative.

\textbf{Limitations and future work.} Liver segmentation retains a canonical-accuracy rotation-robustness trade-off, and higher ligament curvature is associated with lower Dice for both EAMS and VN+EAMS (Appendix~\ref{sec:liver-failures}). Robustness of anatomy-relative coordinates depends on consistent PCA frame selection (Section~\ref{sec:frame-selection}): liver coordinates and predictions are stable in the tested rotations, while dental upper-tail sensitivity motivates resolving the remaining sign choice (Appendix~\ref{sec:pca-sensitivity}). The augmented models incur runtime overhead, and the targeted pose and component diagnostics cover only one fold. Richer intrinsic and anatomy-relative descriptors and additional anatomical tasks remain directions for further evaluation.

\begin{ack}
The author acknowledges funding from the Vector Institute as part of the research internship program. The research was enabled in part by the use of computing resources provided by the Digital Research Alliance of Canada (\texttt{alliancecan.ca}).
\end{ack}

\medskip

{
\small
\bibliography{neurips_2026}
\bibliographystyle{unsrt}
}


\appendix

\section{Data processing}
\label{sec:app-data}
\begin{figure}
    \centering
    \begin{subfigure}{\linewidth}
        \centering
        \includegraphics[width=0.31\linewidth]{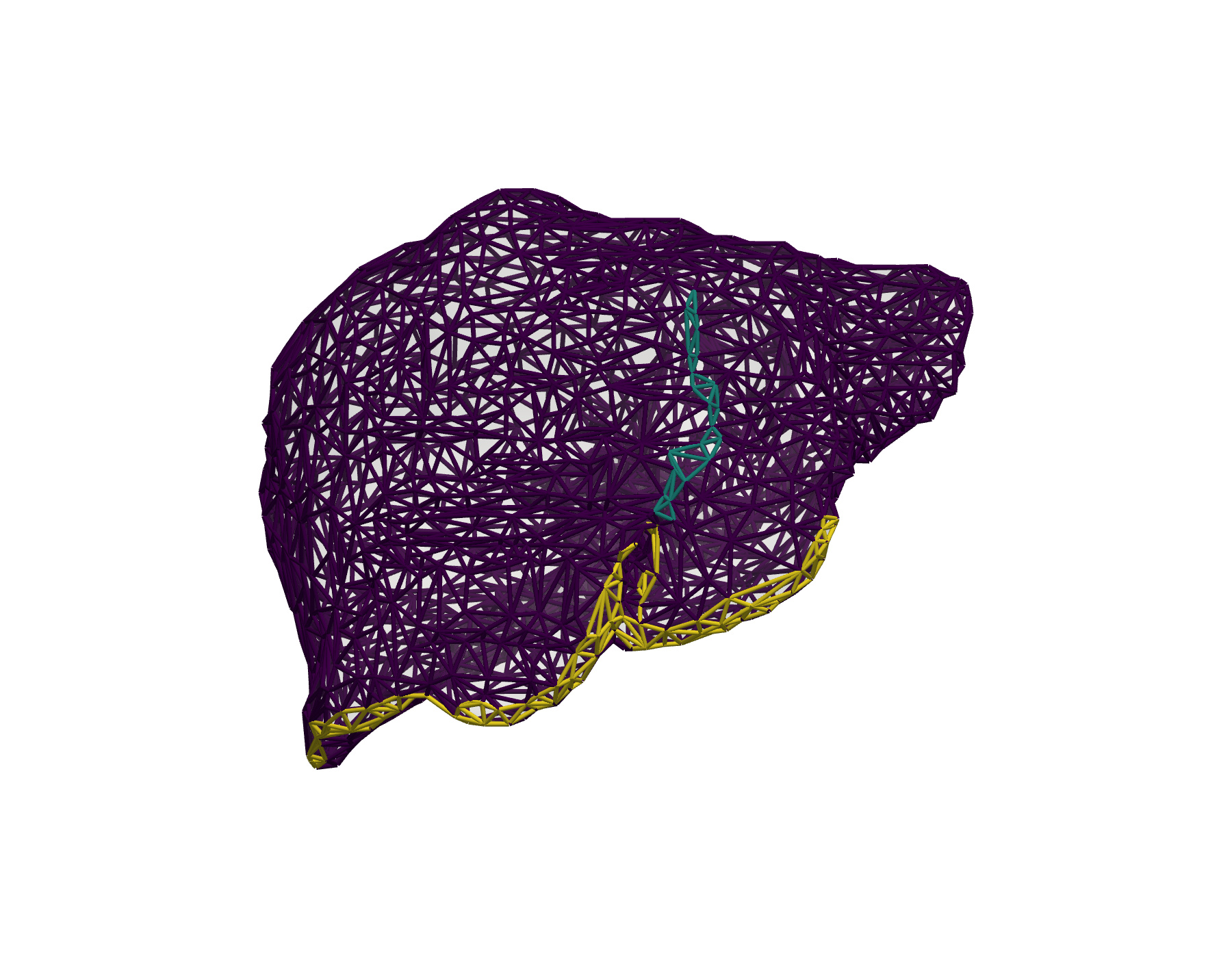}\hfill
        \includegraphics[width=0.31\linewidth]{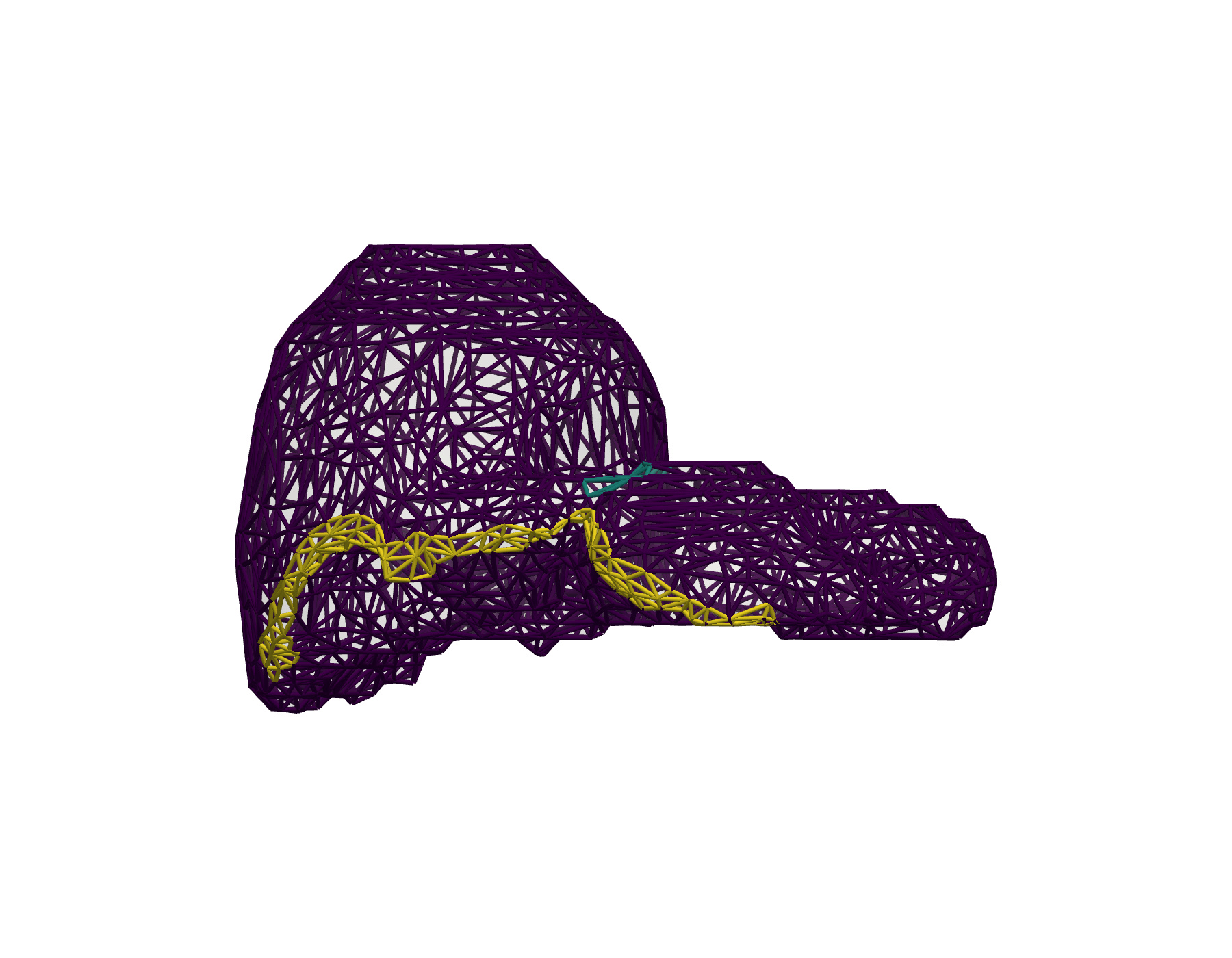}\hfill
        \includegraphics[width=0.31\linewidth]{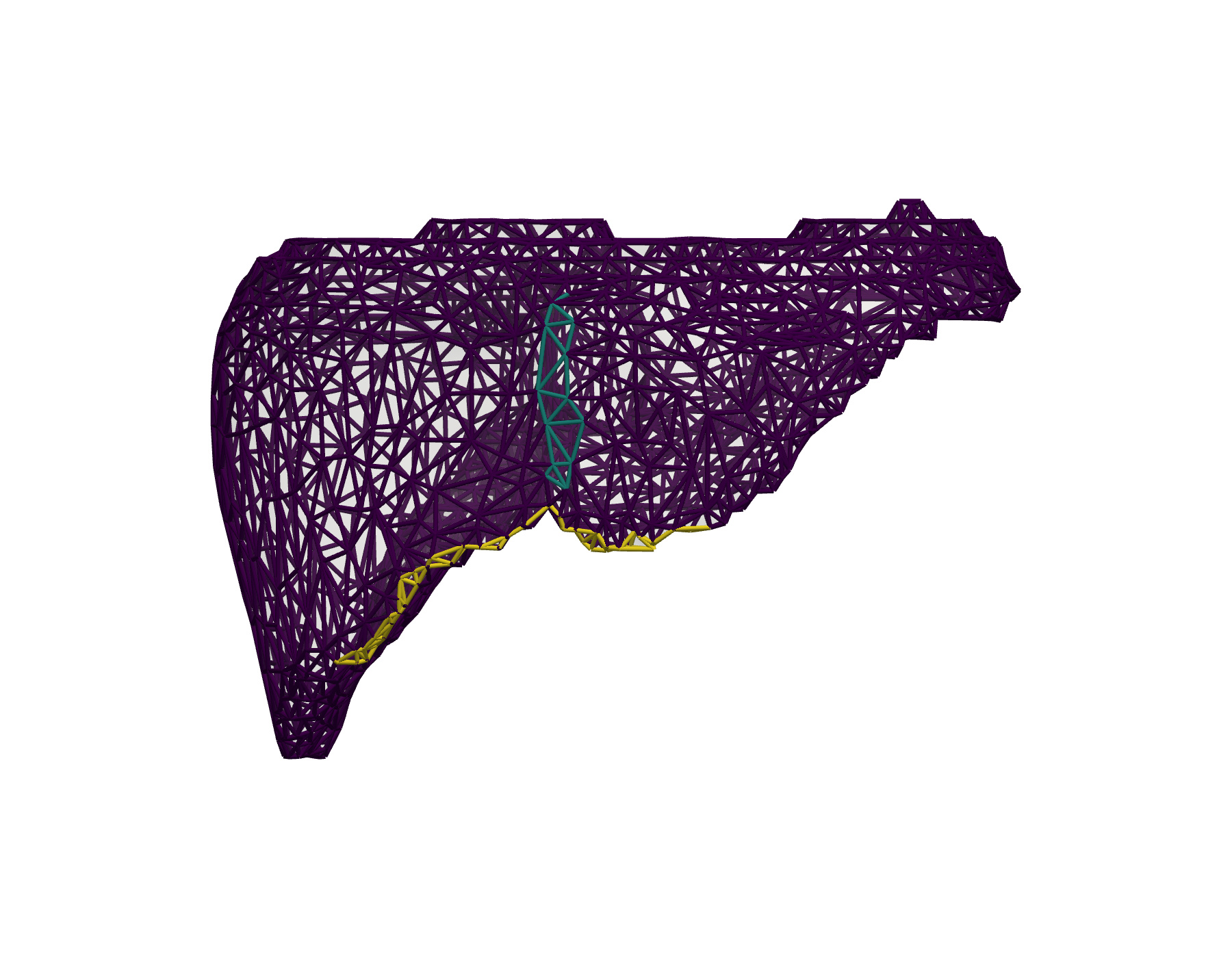}
        \caption{Liver surfaces (3Dircadb, LiTS, Amos). Edge colours denote the falciform ligament (blue) and liver ridge (yellow) annotation regions.}
        \label{fig:liver-examples}
    \end{subfigure}
    \par\medskip
    \begin{subfigure}{\linewidth}
        \centering
        \includegraphics[width=0.23\linewidth]{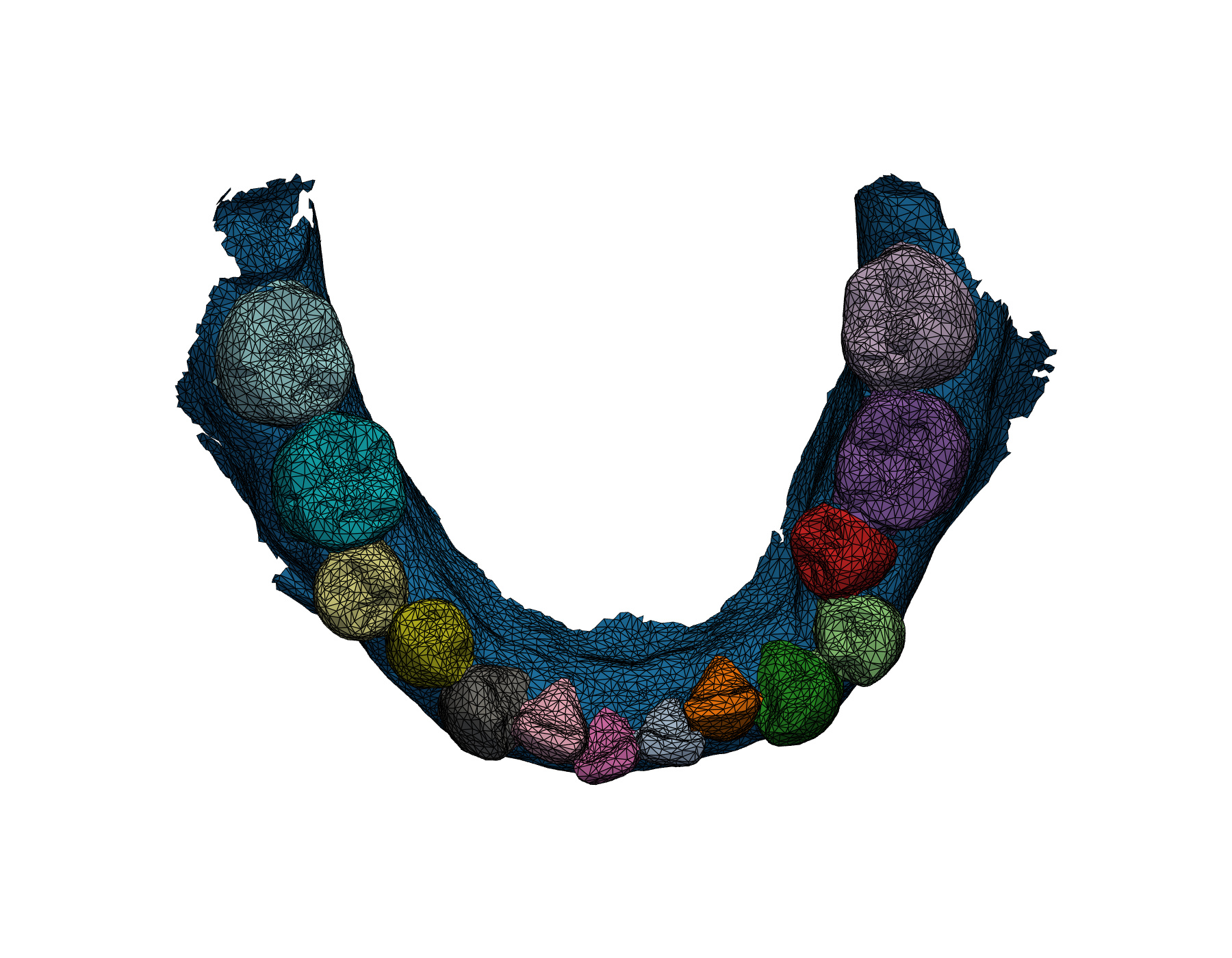}\hfill
        \includegraphics[width=0.23\linewidth]{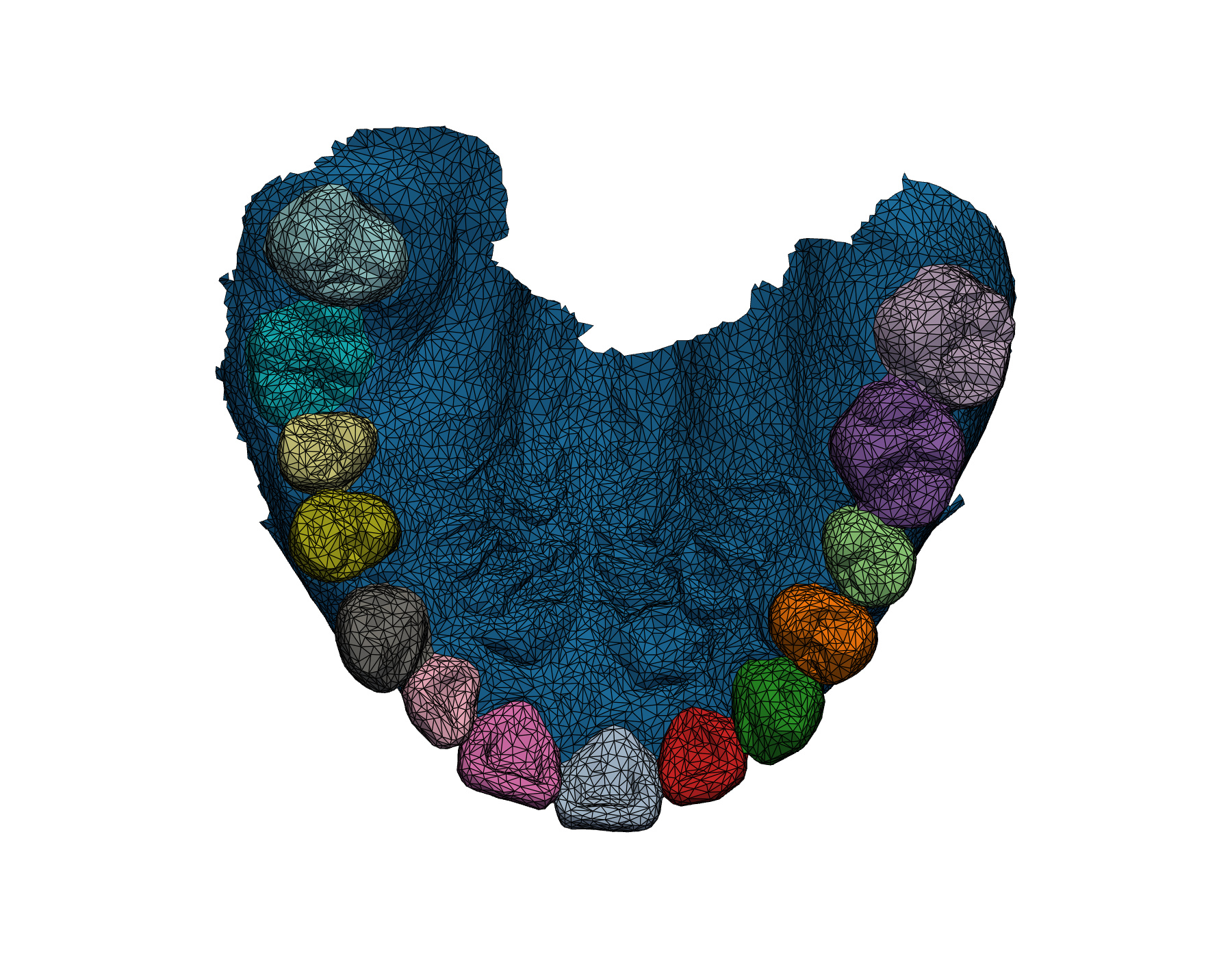}\hfill
        \includegraphics[width=0.23\linewidth]{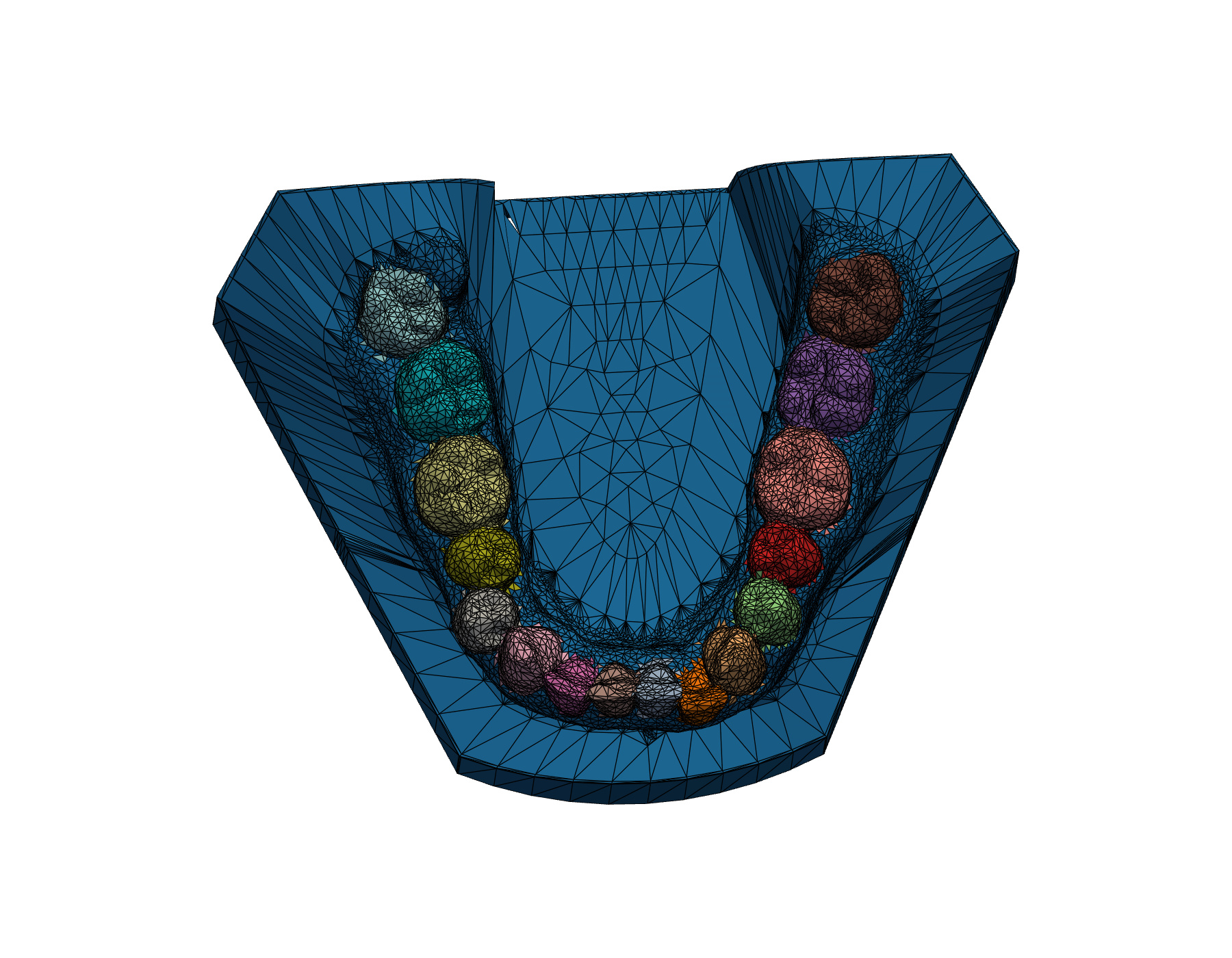}\hfill
        \includegraphics[width=0.23\linewidth]{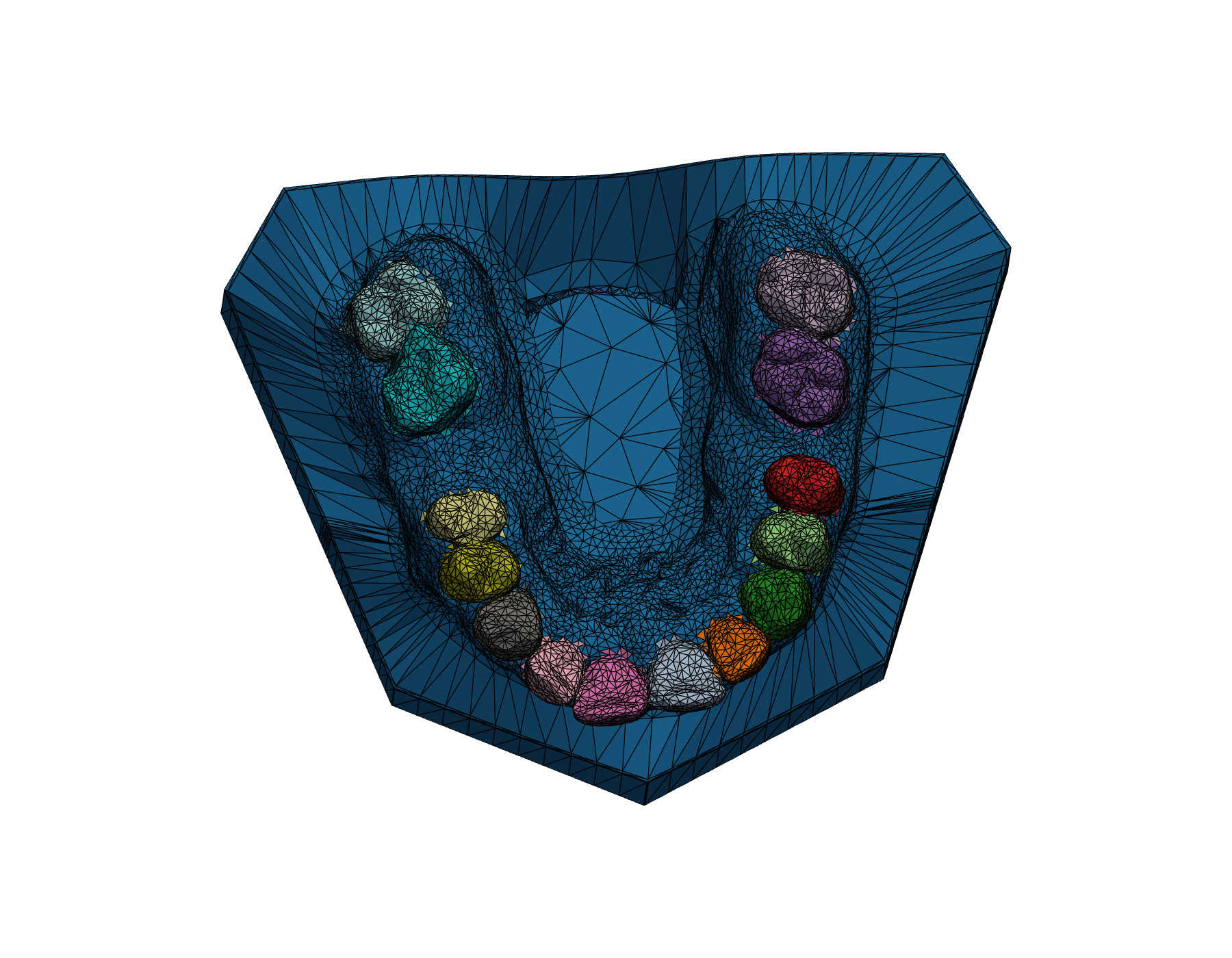}
        \caption{Intraoral scans. Left pair: 3D-IOSSeg (direct scan, richer oral tissue). Right pair: 3DTeethSeg (indirect scan via plaster base).}
        \label{fig:ios-examples}
    \end{subfigure}
    \par\medskip
    \begin{subfigure}{\linewidth}
        \centering
        \includegraphics[width=0.31\linewidth]{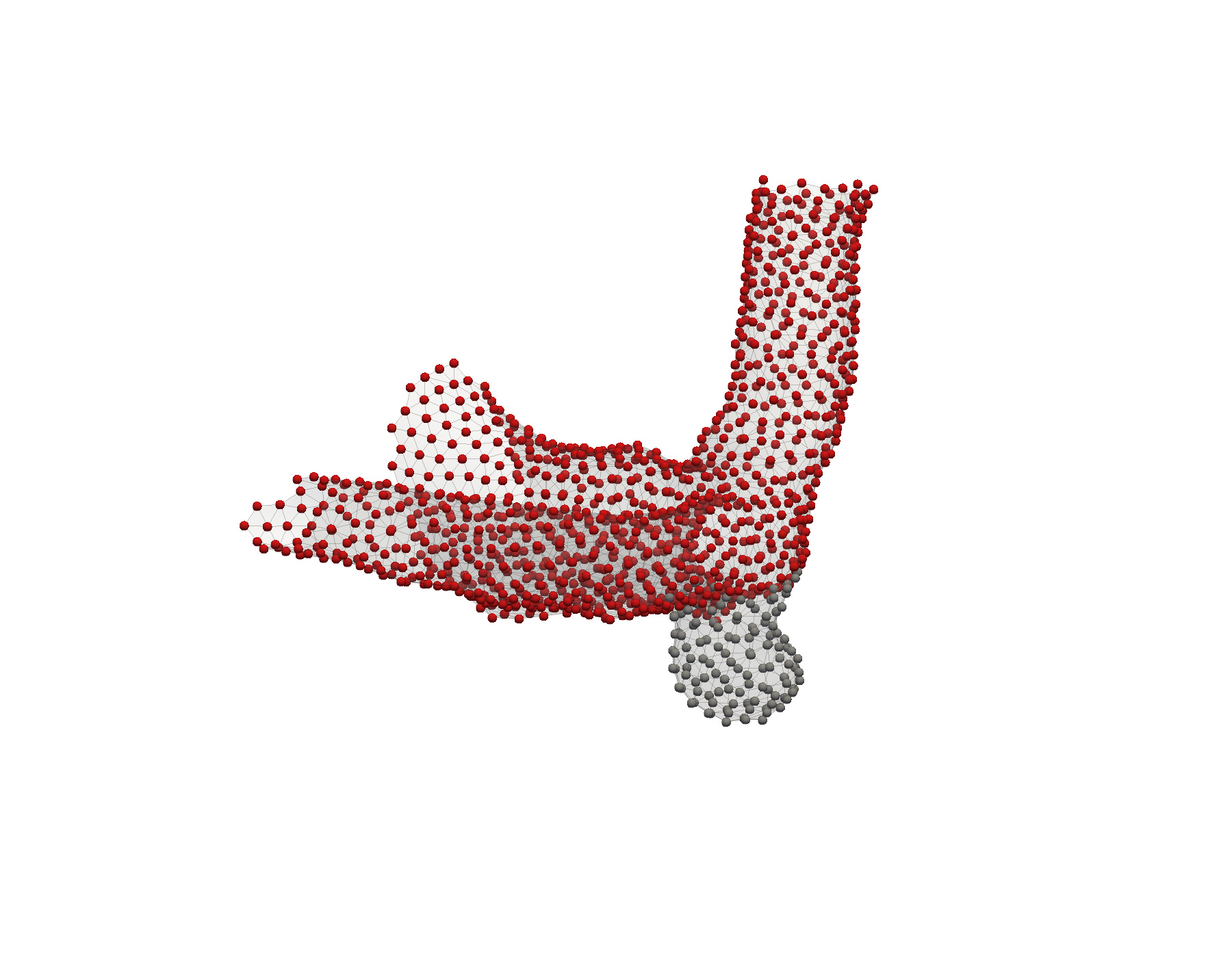}\hfill
        \includegraphics[width=0.31\linewidth]{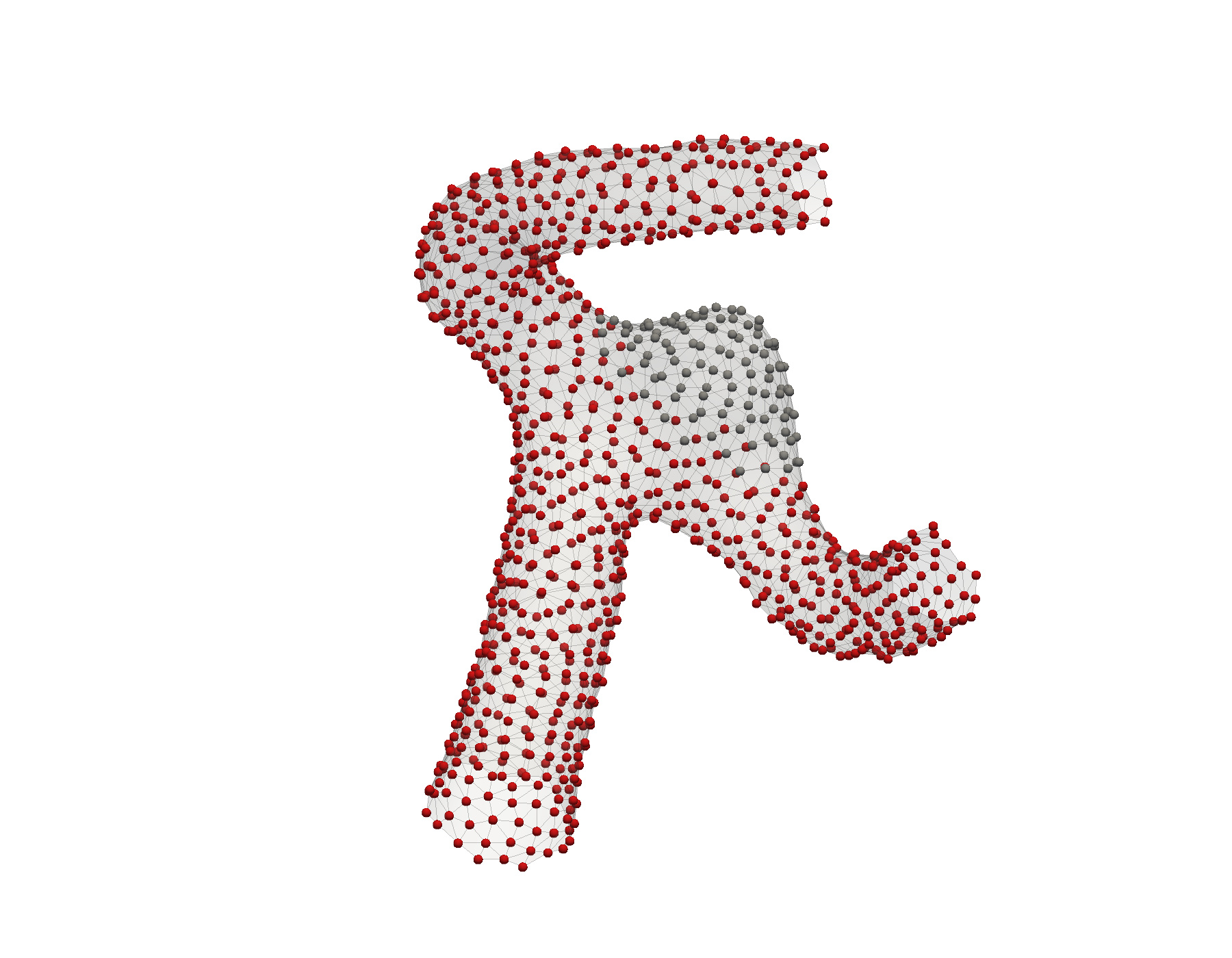}\hfill
        \includegraphics[width=0.31\linewidth]{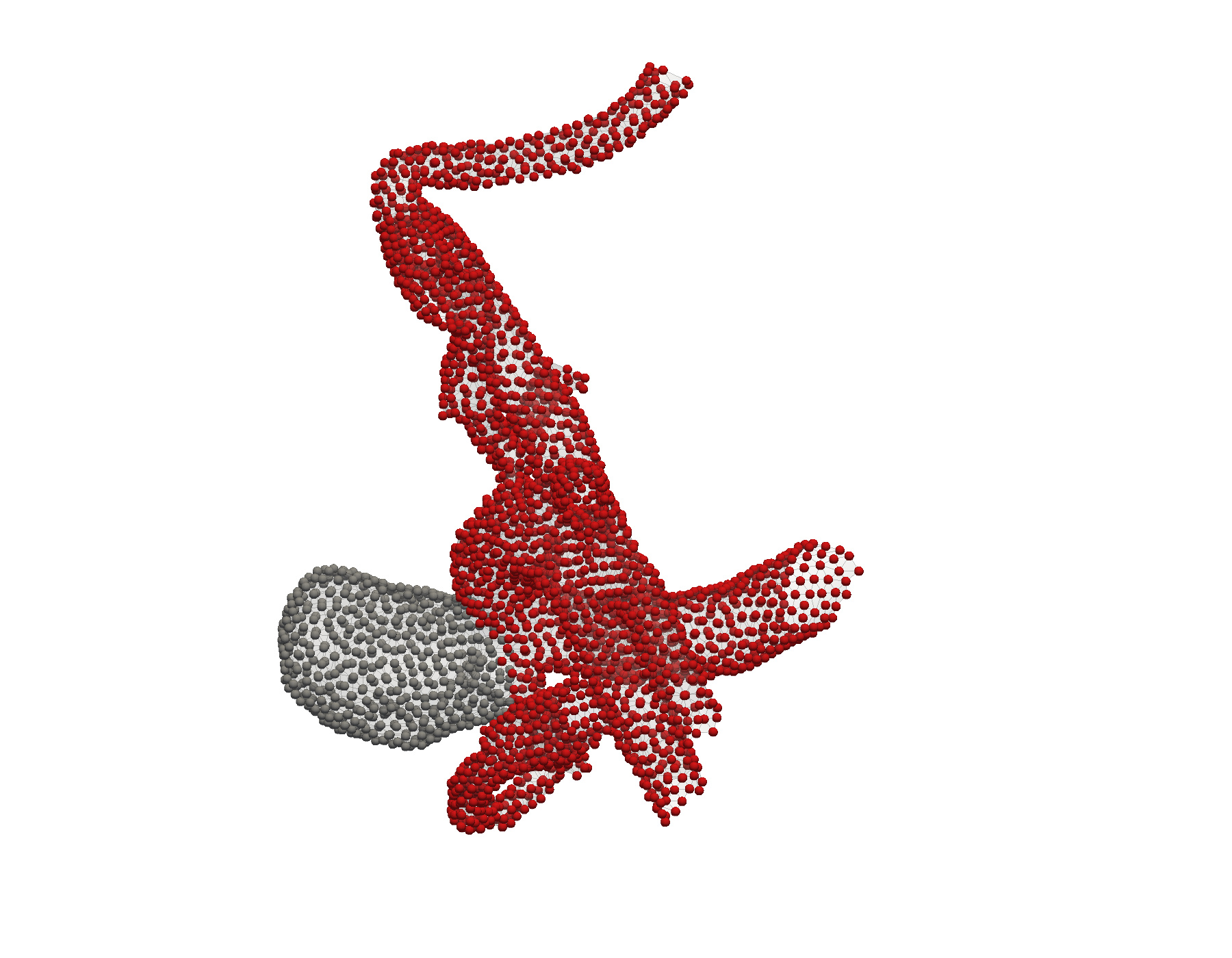}
        \caption{Intracranial aneurysm surfaces from IntrA. Vertex colours indicate aneurysm vs.\ vessel wall labels.}
        \label{fig:intra-examples}
    \end{subfigure}
    \caption{Representative ground-truth segmentations from the three benchmark datasets used in our evaluation.}
    \label{fig:datasets}
\end{figure}

\paragraph{Pipeline overview.}
The data pipeline separates three concerns: raw mesh ingestion, reusable geometric preprocessing, and late feature assembly.
Meshes are first converted into graph-structured objects with their supervision targets.
Expensive geometric quantities---surface normals, adjacency structures, and the Laplace--Beltrami spectral basis---are computed once and stored in per-mesh caches.
Model input tensors are assembled only after batching, so the same cached geometry can support multiple feature configurations without rerunning the full pipeline.
The per-dataset feature configurations are listed in Appendix~\ref{sec:app-features}; the pre-transforms applied before training are described in Appendix~\ref{sec:app-training}.

\paragraph{Mesh cleanup.}
Before any geometric operator is computed, each mesh undergoes a defensive cleanup pass\footnote[1]{This cleanup is the main source of discrepancy between our liver meshes and those in MeshGraphCNN \cite{zhang2025nested}, as seen in Figure~\ref{fig:liver-qual}.}.
Non-finite vertex coordinates are discarded; faces that reference any removed vertex are also removed.
Degenerate triangles (those with repeated vertex indices) and near-zero-area faces are filtered out, and any vertices left unreferenced after face removal are pruned.
Connectivity is then rebuilt on the cleaned mesh.
This pass is not merely cosmetic: the discrete stiffness and mass matrices used for the Laplace--Beltrami eigensystem are sensitive to degenerate elements, and zero-area faces or isolated vertices can produce near-singular Laplacians that destabilise the spectral decomposition.
Cleanup therefore acts as a numerical regularization pass on the mesh geometry itself.
Crucially, it does not perform remeshing or vertex merging, so label alignment between the original annotation and the processed mesh is preserved.

\paragraph{Coordinate normalisation.}
After cleanup, vertex coordinates $\vx_i \in \mathbb{R}^3$ are centred and isotropically scaled:
\begin{equation}
    \vx_i^{\mathrm{norm}} = \frac{\vx_i - \bar{\vx}}{s + \varepsilon},
    \qquad
    \bar{\vx} = \frac{1}{N}\sum_{i=1}^{N} \vx_i,
\end{equation}
where $s$ is the overall mesh extent and $\varepsilon$ is a small constant for numerical stability.
Normalisation makes meshes approximately comparable in magnitude, reduces sensitivity to absolute acquisition scale, and stabilises both the downstream spectral decomposition and network optimisation.

\paragraph{Spectral preprocessing and HKS caching.}
The most expensive per-mesh step is the spectral decomposition of the discrete Laplace--Beltrami operator.
On the normalised mesh, the pipeline solves the generalised eigenproblem
\begin{equation}
    \mW \phi_k = \lambda_k \mM \phi_k,
\end{equation}
where $\mW$ and $\mM$ are the discrete stiffness and mass matrices respectively.
The resulting eigenpairs $\{(\lambda_k, \phi_k)\}$ are stored in the per-mesh cache.
Heat Kernel Signatures~\citep{sun2009concise} are then evaluated from this cached basis as
\begin{equation}
    \mathrm{HKS}(i,t) = \sum_{k} e^{-\lambda_k t}\, \phi_k(i)^2,
\end{equation}
at logarithmically spaced time scales, after which a fixed subset of channels is selected and per-mesh normalised to emphasise relative surface variation.
Caching the eigensystem avoids repeating the spectral decomposition on every training run; the lightweight HKS channel assembly is deferred to the feature stage.

\paragraph{Late feature assembly.}
Node and edge features are assembled from the cached mesh objects only after batching.
Denoting the cached mesh as $\mathcal{M}$, the model inputs are
\begin{equation}
    \mX = f_{\mathrm{node}}(\mathcal{M}), \qquad \mE = f_{\mathrm{edge}}(\mathcal{M}),
\end{equation}
where $f_{\mathrm{node}}$ and $f_{\mathrm{edge}}$ produce the node- and edge-feature matrices respectively.
Because $\mathcal{M}$ is independent of the feature functions, the same cached dataset supports multiple feature configurations without rerunning preprocessing.
Bounded stochastic rotations are applied under the matched training protocol for all main-table methods (Section~\ref{sec:experiments}). Geometry-dependent frame features are recomputed after transformation; Section~\ref{sec:frame-selection} explains the conditions for consistent anatomical coordinates.

\paragraph{Dataset-specific annotation formats.}
The four benchmark datasets share the same geometric preprocessing pipeline but differ in how their raw annotations are stored, requiring dataset-specific conversion to per-vertex or per-edge supervision targets.
\begin{wrapfigure}{r}{0.28\linewidth}
    \centering
    \vspace{-0.5\baselineskip}
    \includegraphics[width=\linewidth]{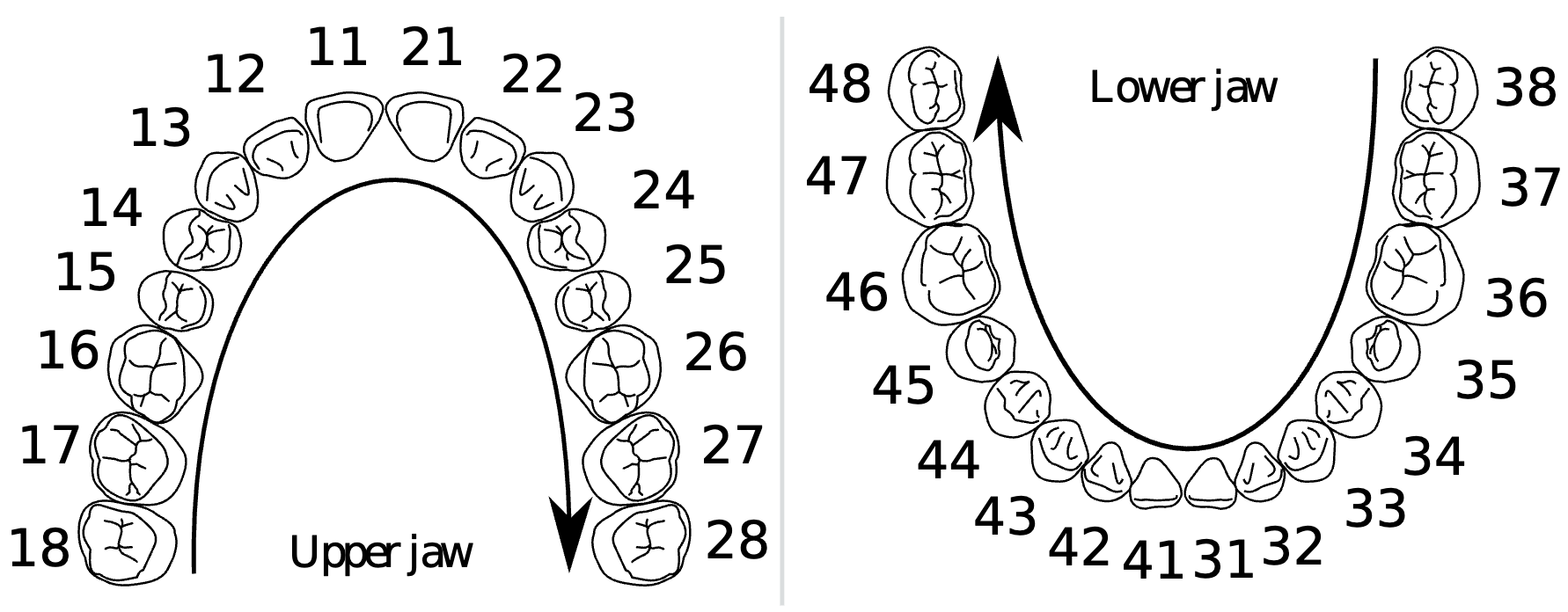}
    \caption{FDI World Dental Federation numbering system for tooth labelling, adapted from \cite{ben20233dteethseg} under CC BY-SA 4.0.}
    \label{fig:ios-notation}
    \vspace{-0.5\baselineskip}
\end{wrapfigure}

\textit{Intraoral scans (Teeth3DS and 3D-IOSSeg).}
Teeth3DS provides per-vertex tooth labels directly in the FDI World Dental Federation numbering system (Figure~\ref{fig:ios-notation}).
3D-IOSSeg instead encodes class information as face colours; face classes are decoded from colour values and per-vertex labels are obtained by majority vote over each vertex's incident faces.
Following the protocol of \cite{xi20253d}, meshes for both dental datasets are downsampled to 16{,}000 faces, and tooth labels are projected back onto the simplified geometry after downsampling.

\textit{Liver surfaces.}
The liver dataset~\citep{zhang2025nested} provides supervision at the edge level.
An external edge graph is supplied alongside each mesh, with per-edge labels covering three anatomical classes: background, falciform ligament, and liver ridge.
When geodesic edge weights are used, they follow an inverse-distance scheme $w_{ij} \propto (d_{ij} + \varepsilon)^{-1}$, so geodesically nearby neighbours receive larger influence during message passing.

\textit{Intracranial aneurysms (IntrA).}
IntrA~\citep{yang2020intra} provides per-vertex binary labels separating the aneurysm sac from the surrounding vessel wall.
When the annotation row count and the mesh vertex count do not match exactly, a nearest-neighbour remapping assigns each mesh vertex the label of its closest annotated point.

\section{Mesh featurization}
\label{sec:app-features}
Table~\ref{tab:features-detail} lists all features used in our experiments with their dimensionalities and computation methods.
Table~\ref{tab:features-per-dataset} summarises the per-dataset configuration.

\begin{table}[h]
\caption{Feature definitions. All features are computed after PyG batching. Node features are concatenated into $\vh_i^{(0)}$; edge features into $\ve_{ij}$.}
\label{tab:features-detail}
\centering
\tablestyle{6pt}{1.2}
\begin{tabular}{llcp{6.5cm}}
\toprule
Feature & Type & Dim & Description \\
\midrule
\texttt{pointwise\_area} & node & 1 & Mean area of incident triangles per vertex, computed via scatter aggregation over the vertex-to-face mapping. \\
\texttt{hks} & node & 8 & Heat Kernel Signature~\citep{sun2009concise} computed from the Laplace--Beltrami spectrum via \texttt{trimesh}; 8 normalised time scales are concatenated. \\
\texttt{dental\_frame\_cylindrical} & node & 3 & Cylindrical coordinates $(r,\theta,z)$ of each vertex in a PCA-derived anatomical dental frame. The mesh is centred, PCA axes are estimated, the anterior-posterior sign is resolved by a quadratic fit, and coordinates are converted to cylindrical form. Used for Teeth3DS and 3D-IOSSeg only. \\
\texttt{canonical\_frame\_cylindrical} & node & 3 & Cylindrical coordinates $(r,\theta,z)$ of each vertex in a PCA-derived anatomical liver frame. The mesh is centred using an area-weighted centre of mass, PCA axes are estimated from an area-weighted covariance, two axis signs are resolved using weighted asymmetry and deterministic fallbacks, and the third axis is derived by a cross product, and coordinates are converted to cylindrical form. Used for liver only. \\
\texttt{com\_fps\_anchor\_distances} & node & 4 & Distances from each node to the centroid and three furthest nodes, computed via a single-pass Furthest Point Sampling (FPS) algorithm. \\
\texttt{degree\_weight} & edge & 1 & Reciprocal of the target-node degree for each directed edge. \\
\texttt{copy\_weight} & edge & 1 & Pre-computed edge weights from the data pipeline when available; falls back to a small constant otherwise. \\
\texttt{dihedrals} & edge & 1 & Dihedral angle between the normals of the two face pairs sharing the edge; defaults to $\pi$ for boundary edges with no adjacent face pair. \\
\bottomrule
\end{tabular}
\end{table}

\begin{table}[h]
\caption{Per-dataset default feature configurations and input dimensionalities. The coordinate channels column counts the number of $\mathbb{R}^3$ vector channels propagated alongside the scalar node features.}
\label{tab:features-per-dataset}
\tablestyle{6pt}{1.2}
\adjustbox{max width=\columnwidth}{%
\begin{tabular}{lllccc}
\toprule
Dataset & Node features & Edge features & Node dim & Edge dim & Coord.\ channels \\
\midrule
All datasets & \texttt{area}, \texttt{hks} & \texttt{deg}, \texttt{copy}, \texttt{dihedrals} & 9 & 3 & 2 (pos.\ + normal) \\
\midrule
\raisebox{3ex}{Liver} &  \shortstack[l]{\texttt{canonical\_cyl} \\ \texttt{canonical\_cart} \\ \texttt{com\_fps\_anchor\_distances}} & \raisebox{3ex}{---} & \raisebox{3ex}{19} & \raisebox{3ex}{3} & \raisebox{3ex}{2} \\
Teeth3DS & \texttt{dental\_cyl} & --- & 12 & 3 & 2 \\
3D-IOSSeg & \texttt{dental\_cyl} & --- & 12 & 3 & 2 \\
IntrA &  --- & --- & 9 & 3 & 2 \\
\bottomrule
\end{tabular}%
}
\end{table}

\begin{algorithm}[ht]
\caption{Dental cylindrical frame}
\label{alg:dental-frame}
\begin{algorithmic}[1]
\Require Vertex coordinates $\{\vx_i\}_{i \in \mathcal{V}}$
\State Compute centroid $\vc \leftarrow \frac{1}{|\mathcal{V}|}\sum_i \vx_i$; centre vertices $\vx_i \leftarrow \vx_i - \vc$
\State Form covariance $\mS \leftarrow \sum_i \vx_i \vx_i^\top$ and compute PCA: $[\mU, \mLambda] \leftarrow \mathrm{eig}(\mS)$
\State Order PCA axes consistently and construct a right-handed frame
\State \textbf{Resolve AP/$Y$ sign:} use a quadratic fit to the projected dental arch to orient the $Y$ axis by concavity
\State Re-orthonormalise to preserve right-handedness; the simultaneous $(X,Z)\mapsto(-X,-Z)$ choice remains unresolved
\State Rotate into dental frame: $\vy_i \leftarrow \mU^\top \vx_i$
\State Shift posterior baseline: $[\vy_i]_2 \leftarrow [\vy_i]_2 - \min_j [\vy_j]_2$
\State Convert to cylindrical: $(r_i, \theta_i, z_i) \leftarrow \mathrm{cart2cyl}(\vy_i)$
\Ensure Per-vertex features $(r_i, \theta_i, z_i)$
\end{algorithmic}
\end{algorithm}

\paragraph{Transformation of the dental frame.}
Under $\vx_i \mapsto \mR\vx_i+\vt$ with $\mR\in\mathrm{SO}(3)$, centred coordinates transform as $\mR(\vx_i-\vc)$ and covariance as $\mS\mapsto\mR\mS\mR^\top$.
For distinct eigenvalues with fixed ordering, PCA determines $\mU\mapsto\mR\mU\mD$ up to a diagonal sign matrix $\mD$.
Thus, PCA supplies a rotation-equivariant set of candidate frames; coordinates in a selected frame are invariant only if sign selection is consistent.
The arch-concavity rule selects the AP/$Y$ direction and right-handedness constrains the remaining signs, but the simultaneous $(X,Z)\mapsto(-X,-Z)$ choice remains.
Repeated eigenvalues permit a continuous choice within an eigenspace, while near-repeated eigenvalues can amplify numerical perturbations.
Section~\ref{sec:frame-selection} explains how these conditions relate to the small fixed-angle score variation; Appendix~\ref{sec:pca-sensitivity} reports coordinate and prediction sensitivity.
Reflections are outside the dental target group because left/right tooth identity is label-relevant.

\paragraph{Liver frame: area-weighted PCA.}
The liver uses the same overall PCA-to-cylindrical construction as the dental frame, but replaces the uniform centroid and covariance with area-weighted quantities to reduce sensitivity to irregular tessellation.
Specifically, if $A_i$ denotes the local vertex area associated with vertex $i$, we compute the area-weighted centre of mass
\begin{equation}
\vmu = \frac{1}{\sum_{i=1}^{N} A_i} \sum_{i=1}^{N} A_i \vx_i
\end{equation}
and the area-weighted covariance
\begin{equation}
\mC = \frac{1}{\sum_{i=1}^{N} A_i} \sum_{i=1}^{N} A_i (\vx_i - \vmu)(\vx_i - \vmu)^\top.
\end{equation}
This change matters more for liver than for teeth because liver meshes exhibit much larger inter-patient variation in lobe shape and are also more vulnerable to local remeshing artifacts near vessels or thin folds.
Weighting by area makes the canonical frame depend on the organ's physical surface distribution rather than on raw vertex density: densely triangulated nuisance regions contribute little because their associated $A_i$ are small, whereas the dominant right-lobe versus tapering left-lobe mass distribution remains stable.
In practice, this yields a more anatomically faithful centre and principal axes, making the liver cylindrical coordinates less sensitive to mesh-resolution artifacts subject to the frame-selection conditions above. The liver construction resolves two axis signs from weighted asymmetry/third-moment criteria with deterministic fallbacks, then derives the third by a cross product; the empirical sensitivity analysis is in Appendix~\ref{sec:pca-sensitivity}.

\begin{algorithm}[ht]
\caption{Soft regional aggregator (SRA) update}
\label{alg:sr-update}
\begin{algorithmic}[1]
\Require Node features $\{\vh_i\}$, coordinates $\{\vx_i\}$, number of regions $K$, learned residual weight $\alpha$
\For{each node $i$}
    \State $\va_i \leftarrow \mathrm{softmax}(\phi_a(\vh_i)) \in \mathbb{R}^K$ \quad (optionally augmented with dist. to CoM $\vu_i = \|\vx_i - \bar \vx\|$)
\EndFor
\State Stack into assignment matrix: $[\mA]_{ik} \leftarrow [\va_i]_k$, so $\mA \in \mathbb{R}^{N \times K}$
\For{each region $k = 1, \ldots, K$}
    \State $\vr_k \leftarrow \sum_i [\mA]_{ik}\, \vh_i$ \quad (optionally augmented with geom. feature $\vu^r_k = \sum_i [\mA]_{ik}\, \vu_i$)
\EndFor
\State Mix region tokens: $[\hat{\vr}_1, \ldots, \hat{\vr}_K] \leftarrow \mathrm{TransformerEncoder}([\vr_1, \ldots, \vr_K])$
\For{each node $i$}
    \State $\vh_i \leftarrow \vh_i + \alpha\, \sum_k [\mA]_{ik}\, \phi_{\mathrm{proj}}(\hat{\vr}_k)$
\EndFor
\Ensure Updated node features $\{\vh_i\}$
\end{algorithmic}
\end{algorithm}

\begin{algorithm}[ht]
\caption{Virtual node update (per EMNN layer)}
\label{alg:fast-vn}
\begin{algorithmic}[1]
\Require Node features $\{\vh_i\}$, coordinates $\{\vx_i\}$, mesh edges $E$, faces $\mathcal{F}$; virtual features $\{\vv_k\}_{k=1}^{V}$, virtual coordinates $\{\vu_k\}$
\State Compute standard EMNN edge messages $\vm_i^{\mathrm{edge}}$ and coordinate updates $\Delta \vx_i^{\mathrm{edge}}$
\State Compute face-area messages $\vm_i^{\mathrm{face}}$ and coordinate updates $\Delta \vx_i^{\mathrm{face}}$
\State For each graph, compute the mean node coordinate $\bar{\vx}$.
\State $\vm^v_k \leftarrow \|\vu_k - \bar{\vx}\|^2$ \quad (correlation feature for virtual node $k$)
\For{each node $i$ and each virtual node $k$ in its graph}
    \State $\Delta \vx_{i \to k} \leftarrow \vu_k - \vx_i$
    \State $\vm_{ik}^{\mathrm{virt}} \leftarrow \phi_{rv}(\vh_i,\, \vv_k,\, \|\Delta \vx_{i \to k}\|^2,\, \vm_k^v)$
\EndFor
\State Aggregate toward nodes:
\Statex \qquad $\vm_i^{\mathrm{virt}} \leftarrow \frac{1}{V} \sum_{k=1}^{V} \vm_{ik}^{\mathrm{virt}}$
\Statex \qquad $\Delta \vx_i^{\mathrm{virt}} \leftarrow \frac{1}{V} \sum_{k=1}^{V} \phi_{\mathrm{virt}\to\mathrm{node}}(\vm_{ik}^{\mathrm{virt}})(\vx_i - \vu_k)$
\State Aggregate toward virtual nodes:
\Statex \qquad $\bar{\vm}_k^{\mathrm{virt}} \leftarrow \frac{1}{N_g} \sum_{i \in g} \vm_{ik}^{\mathrm{virt}}$
\Statex \qquad $\Delta \vu_k \leftarrow \frac{1}{N_g} \sum_{i \in g} \phi_{\mathrm{node}\to\mathrm{virt}}(\vm_{ik}^{\mathrm{virt}})(\vu_k - \vx_i)$
\State Update virtual states: $\vv_k \leftarrow \vv_k + \phi_v(\vv_k,\, \bar{\vm}_k^{\mathrm{virt}})$
\State Optionally project or collapse coordinate channels before adding the propagated updates
\State Update node coordinates: $\vx_i \leftarrow \vx_i + \Delta \vx_i^{\mathrm{edge}} + \Delta \vx_i^{\mathrm{face}} + \Delta \vx_i^{\mathrm{virt}}$
\State Update virtual coordinates: $\vu_k \leftarrow \vu_k + \Delta \vu_k$
\State Update node features:
\Statex \qquad $\vh_i \leftarrow \phi_h([\vh_i,\, \vm_i^{\mathrm{edge}},\, \vm_i^{\mathrm{face}},\, \vm_i^{\mathrm{virt}}])$
\Ensure Updated $\{\vh_i, \vx_i, \vv_k, \vu_k\}$
\end{algorithmic}
\end{algorithm}

\section{Training and implementation details}
\label{sec:app-training}
\paragraph{Code availability.}
Training code and configurations are publicly available at \coderepourl, including mesh preprocessing and label conversion, HKS-cache generation, PCA-frame construction, split definitions, and example configurations.

\paragraph{Optimiser and learning rate schedule.}
All models are trained with AdamW ($\beta_1 = 0.9$, $\beta_2 = 0.999$, weight decay $0.01$).
We use a plateau-based learning rate schedule: initial LR $10^{-3}$, reduction factor $0.6$, minimum LR $10^{-5}$.
Gradient norms are clipped to $1.0$.
Training runs for 100 epochs with validation every 5 epochs; all inputs are 32-bit floating point.

\paragraph{Batch size and cross-validation.}
All experiments use batch size 1, except liver segmentation which uses batch size 8.
We perform 5-fold cross-validation for the liver, 3D-IOSSeg, and IntrA datasets, and a single train/test split for Teeth3DS following the protocol of \cite{ben20233dteethseg}.

\paragraph{Loss weights.}
The prediction loss coefficient is set to $1.0$ for all tasks.
The contrastive boundary loss coefficient is set to $10.0$ for all tasks, except for liver segmentation where it is set to $1.0$.
For liver segmentation, the local continuity loss uses $\lambda_{\text{cont}} = 1.0$.
For SRA runs, the assignment regularization losses use $\lambda_{\text{div}} = \lambda_{\text{eq}} = 1.0$.
For virtual-node runs, the virtual-node coordinate loss uses $w_{\mathit{vv}} = w_{\mathit{rv}} = 1.0$.

\paragraph{Global-token configuration.}
For the liver experiments, SRA+EAMS uses $K=32$ soft regional aggregators and VN+EAMS uses $V=8$ virtual nodes.

\paragraph{Encoder architecture.}
The encoder is an augmented EMNN with hidden dimensions $(128, 128, 128)$, SiLU activations, and batch normalization. We use a multi-vector channel \cite{levy2023using,trang20243}, 
which replaces vector features $\vx_i \in \mathbb{R}^3$ with $\mX_i \in \mathbb{R}^{3C}$ for some small $C$; we typically set $C=2$ where the extra channel comes from the normal vector.

\paragraph{Decoder architecture.}
For all datasets except liver segmentation, predictions are produced by an MLP decoder applied independently at each node, 
$\hat{\vy}_i = f_{\mathrm{dec}}(\vh_i)$,
where $f_{\mathrm{dec}}$ has hidden dimensions $(128, 128, 128)$, SiLU activations, and dropout $0.1$.
For liver segmentation, supervision is defined on directed edges, so we instead decode edgewise messages.
For an edge $(j,i)$, the decoder forms
\[
\vz_{ji} = [\vh_j \,\|\, \vh_i \,\|\, \ve_{ji} \,\|\, \psi(\|\vx_j - \vx_i\|_2)],
\]
predicts an edge label
\(
\hat{\vy}_{ji} = f_{\mathrm{edge}}(\vz_{ji}),
\)
Here $f_{\mathrm{edge}}$ uses the same hidden dimensions, activation, and dropout as the nodewise MLP decoder, while the distance embedding $\psi$ is a one-hidden-layer MLP with hidden width $128$ and output dimension $16$.

\paragraph{Pre-transforms.}
Meshes are pre-processed once before training with the following deterministic transforms: coordinate normalisation, vertex-normal estimation, face-adjacency structures, vertex-to-edge mapping, and cached HKS computation.
All methods in the main comparison tables use bounded stochastic training rotations, with angles sampled uniformly from $[-0.15\pi,0.15\pi]$, as described in Section~\ref{sec:experiments}. Splits and model-selection protocol are held fixed across the matched comparisons. Deterministic centring removes translation, so separate translation augmentation is unnecessary. Dental and liver PCA coordinates are recomputed from transformed geometry. The random-axis test protocol is described in Appendix~\ref{sec:random-axis}, and the targeted pose and context ablations use a single fold rather than five-fold averages. In the CBL ablation, only the embedding loss is removed; the prediction loss is retained.

\paragraph{Metrics.}
We report Intersection over Union (IoU) across all tasks.
For liver segmentation we additionally report Dice, Chamfer distance \cite{wu2021balanced}, and Hausdorff distance.
For IntrA we additionally report Dice.

\paragraph{Training runtimes.}
All experiments are run on a single H100 GPU. Table~\ref{tab:eams-runtime} reports per-epoch wall-clock runtimes for the base EAMS model.
Relative to EAMS, SRA+EAMS increases runtime by a factor of $1.45\times$ and VN+EAMS by $1.64\times$.

\begin{table}[h]
\caption{Per-epoch runtime for EAMS.}
\label{tab:eams-runtime}
\centering
\tablestyle{6pt}{1.2}
\begin{tabular}{lc}
\toprule
Dataset & Runtime / epoch \\
\midrule
Liver & 7 s \\
3D-IOSSeg & 9.2 s \\
Teeth3DS & 72 s \\
IntrA & 1.2 s \\
\bottomrule
\end{tabular}
\end{table}

\section{Broader impacts}
\label{sec:app-broader-impacts}
This work is intended to support clinical and research workflows that rely on anatomical surface segmentation. More robust mesh segmentation under pose variation could reduce manual annotation burden, improve consistency across acquisition settings, and make downstream geometric analysis or planning tools less sensitive to scanner orientation. These benefits are most plausible in settings where segmentation is used as a decision-support component and where models are validated for the target anatomy, acquisition pipeline, and patient population.

The main risk is harm from incorrect results in intended clinical use. If a segmentation model fails on atypical anatomy, poor-quality meshes, or shifts in acquisition or reconstruction pipelines, it could mislabel clinically relevant structures and propagate errors into downstream measurements, visualisation, or treatment-planning software. For this reason, systems based on this technology should be deployed with clinician oversight, dataset-appropriate validation, and task-specific quality control rather than as fully autonomous tools.

\section{Additional experimental results}
\label{sec:app-results}

\subsection{Intracranial aneurysm segmentation: full robustness results}

Table~\ref{tab:intra-robust-full} extends Table~\ref{tab:intra-main} with IoU scores for both segmentation targets across the fixed-angle rotations and reflection test. EAMS, SRA+EAMS, and VN+EAMS are E(3)-invariant, so their scores are identical to the unperturbed results in Table~\ref{tab:intra-main} and are omitted here.

\begin{table*}[t]
\caption{%
  Full robustness results for intracranial aneurysm segmentation on the IntrA dataset.
  Results are mean $\pm$ std over five cross-validation folds.
  EAMS, SRA+EAMS, and VN+EAMS are E(3)-invariant; their scores are identical to Table~\ref{tab:intra-main} under all listed transformations and are omitted.
}
\label{tab:intra-robust-full}
\tablestyle{5pt}{1.15}
\adjustbox{max width=\textwidth}{%
\begin{tabular}{llcccc}
\toprule
& & \multicolumn{2}{c}{Parent vessel} & \multicolumn{2}{c}{Aneurysm} \\
\cmidrule(lr){3-4}\cmidrule(lr){5-6}
Method & Condition & Dice (\%) $\uparrow$ & IoU (\%) $\uparrow$ & Dice (\%) $\uparrow$ & IoU (\%) $\uparrow$ \\
\midrule
DGCNN \citep{wang2019dynamic}
  & Canonical & \pmval{97.16}{0.59} & \pmval{94.61}{1.06} & \pmval{88.17}{1.69} & \pmval{80.59}{2.58} \\
  & Rot-$z$ 15\textdegree & \pmval{96.91}{0.64} & \pmval{94.14}{1.17} & \pmval{87.45}{1.71} & \pmval{79.30}{2.79} \\
  & Rot-$z$ 40\textdegree & \pmval{95.65}{0.19} & \pmval{91.85}{0.35} & \pmval{81.75}{0.40} & \pmval{71.51}{0.32} \\
  & Refl-$x$ & \pmval{95.85}{0.34} & \pmval{92.21}{0.61} & \pmval{81.28}{2.76} & \pmval{71.61}{3.24} \\
\midrule
PTv3 \citep{wu2024point}
  & Canonical & \pmval{95.28}{1.38} & \pmval{91.26}{2.41} & \pmval{81.29}{5.28} & \pmval{72.30}{6.40} \\
  & Rot-$z$ 15\textdegree & \pmval{95.11}{1.25} & \pmval{90.94}{2.17} & \pmval{80.36}{4.96} & \pmval{71.21}{5.78} \\
  & Rot-$z$ 40\textdegree & \pmval{94.31}{1.05} & \pmval{89.55}{1.80} & \pmval{77.24}{4.26} & \pmval{67.65}{5.01} \\
  & Refl-$x$ & \pmval{93.67}{0.87} & \pmval{88.49}{1.49} & \pmval{74.70}{4.44} & \pmval{64.88}{4.47} \\
\bottomrule
\end{tabular}%
}
\end{table*}

\begin{figure*}[t]
\centering
\makebox[0.245\textwidth][c]{\textbf{Ground truth}}\hfill
\makebox[0.245\textwidth][c]{\textbf{EAMS}}\hfill
\makebox[0.245\textwidth][c]{\textbf{SRA+EAMS}}\hfill
\makebox[0.245\textwidth][c]{\textbf{VN+EAMS}}

\par\smallskip

\begin{subfigure}[t]{0.245\textwidth}
  \qualimgintra{intra/emnn/base/AN54-1_full_edge_labels_xz_gt}
\end{subfigure}\hfill
\begin{subfigure}[t]{0.245\textwidth}
  \qualimgintra{intra/emnn/base/AN54-1_full_edge_labels_xz_pred}
\end{subfigure}\hfill
\begin{subfigure}[t]{0.245\textwidth}
  \qualimgintra{intra/sra_emnn/base/AN54-1_full_edge_labels_xz_pred}
\end{subfigure}\hfill
\begin{subfigure}[t]{0.245\textwidth}
  \qualimgintra{intra/vn_emnn/base/AN54-1_full_edge_labels_xz_pred}
\end{subfigure}

\par\smallskip
\begin{subfigure}[t]{0.245\textwidth}
  \qualimgintra{intra/emnn/base/AN193-1_full_edge_labels_xz_gt}
\end{subfigure}\hfill
\begin{subfigure}[t]{0.245\textwidth}
  \qualimgintra{intra/emnn/base/AN193-1_full_edge_labels_xz_pred}
\end{subfigure}\hfill
\begin{subfigure}[t]{0.245\textwidth}
  \qualimgintra{intra/sra_emnn/base/AN193-1_full_edge_labels_xz_pred}
\end{subfigure}\hfill
\begin{subfigure}[t]{0.245\textwidth}
  \qualimgintra{intra/vn_emnn/base/AN193-1_full_edge_labels_xz_pred}
\end{subfigure}

\par\smallskip
\begin{subfigure}[t]{0.245\textwidth}
  \qualimgintra{intra/emnn/base/AN195_full_edge_labels_xz_gt}
\end{subfigure}\hfill
\begin{subfigure}[t]{0.245\textwidth}
  \qualimgintra{intra/emnn/base/AN195_full_edge_labels_xz_pred}
\end{subfigure}\hfill
\begin{subfigure}[t]{0.245\textwidth}
  \qualimgintra{intra/sra_emnn/base/AN195_full_edge_labels_xz_pred}
\end{subfigure}\hfill
\begin{subfigure}[t]{0.245\textwidth}
  \qualimgintra{intra/vn_emnn/base/AN195_full_edge_labels_xz_pred}
\end{subfigure}

\par\medskip
\hrule
\par\smallskip

\begin{subfigure}[t]{0.245\textwidth}
  \qualimgintra{intra/emnn/rotate_40/AN125_full_edge_labels_xz_gt}
\end{subfigure}\hfill
\begin{subfigure}[t]{0.245\textwidth}
  \qualimgintra{intra/emnn/rotate_40/AN125_full_edge_labels_xz_pred}
\end{subfigure}\hfill
\begin{subfigure}[t]{0.245\textwidth}
  \qualimgintra{intra/sra_emnn/rotate_40/AN125_full_edge_labels_xz_pred}
\end{subfigure}\hfill
\begin{subfigure}[t]{0.245\textwidth}
  \qualimgintra{intra/vn_emnn/rotate_40/AN125_full_edge_labels_xz_pred}
\end{subfigure}

\par\smallskip
\begin{subfigure}[t]{0.245\textwidth}
  \qualimgintra{intra/emnn/rotate_40/AN178_full_edge_labels_xz_gt}
\end{subfigure}\hfill
\begin{subfigure}[t]{0.245\textwidth}
  \qualimgintra{intra/emnn/rotate_40/AN178_full_edge_labels_xz_pred}
\end{subfigure}\hfill
\begin{subfigure}[t]{0.245\textwidth}
  \qualimgintra{intra/sra_emnn/rotate_40/AN178_full_edge_labels_xz_pred}
\end{subfigure}\hfill
\begin{subfigure}[t]{0.245\textwidth}
  \qualimgintra{intra/vn_emnn/rotate_40/AN178_full_edge_labels_xz_pred}
\end{subfigure}

\par\smallskip
\begin{subfigure}[t]{0.245\textwidth}
  \qualimgintra{intra/emnn/rotate_40/AN196-2_full_edge_labels_xz_gt}
\end{subfigure}\hfill
\begin{subfigure}[t]{0.245\textwidth}
  \qualimgintra{intra/emnn/rotate_40/AN196-2_full_edge_labels_xz_pred}
\end{subfigure}\hfill
\begin{subfigure}[t]{0.245\textwidth}
  \qualimgintra{intra/sra_emnn/rotate_40/AN196-2_full_edge_labels_xz_pred}
\end{subfigure}\hfill
\begin{subfigure}[t]{0.245\textwidth}
  \qualimgintra{intra/vn_emnn/rotate_40/AN196-2_full_edge_labels_xz_pred}
\end{subfigure}

\caption{%
  Additional qualitative IntrA comparisons for the invariant mesh variants on the cases in Figure~\ref{fig:intra-qual}. The top half shows canonical examples; the bottom half shows a separate set after a 40\textdegree{} $z$-axis rotation.
  This appendix-only figure restores the omitted EAMS and SRA+EAMS predictions, showing that all three mesh-based variants remain visually stable under 40\textdegree{} rotation, with SRA+EAMS and VN+EAMS typically producing the cleanest aneurysm boundaries.
}
\label{fig:intra-qual-ours}
\end{figure*}

\subsection{Intraoral scan tooth segmentation: per-class results}

Tables~\ref{tab:iosseg-pertooth-left}--\ref{tab:teeth3ds-pertooth-right} extend
the aggregate comparison with per-class IoU (\%) for the left and right jaw halves of both intraoral benchmarks.
These appendix rows are class-wise summaries, whereas the main-table Average IoU averages each mesh over its ground-truth-present classes and then averages over meshes.
Column headers use squashed FDI notation: Txy/x$'$y$'$ averages the mirrored upper- and lower-jaw classes.
For 3D-IOSSeg (Tables~\ref{tab:iosseg-pertooth-left}--\ref{tab:iosseg-pertooth-right}), all methods use the matched training augmentation of Table~\ref{tab:iosseg-main}; EAMS methods are listed at Baseline only, with their rotation variation summarized in that table. Fast-TGCN's augmentation protocol is unchanged.
For Teeth3DS (Tables~\ref{tab:teeth3ds-pertooth-left}--\ref{tab:teeth3ds-pertooth-right}), all conditions are listed for all methods to report the frame-dependent variation explained in Section~\ref{sec:frame-selection}; no standard deviation is reported (single fold).

\begin{table*}[t]
\caption{%
  Per-class IoU (\%) for the left jaw half on 3D-IOSSeg \citep{li2024fine} (Gingiva and T11/31--T18/38).
  Results are mean $\pm$ std over five cross-validation folds under matched training augmentation, as in Table~\ref{tab:iosseg-main}. EAMS-family rotation variation is discussed in Section~\ref{sec:frame-selection}.
}
\label{tab:iosseg-pertooth-left}
\tablestyle{4pt}{1.1}
\adjustbox{max width=\textwidth}{%
\begin{tabular}{llccccccccc}
\toprule
Method & Condition
  & Gingiva $\uparrow$ & T11/31 $\uparrow$ & T12/32 $\uparrow$ & T13/33 $\uparrow$
  & T14/34 $\uparrow$ & T15/35 $\uparrow$ & T16/36 $\uparrow$ & T17/37 $\uparrow$ & T18/38 $\uparrow$ \\
\midrule
DGCNN \citep{wang2019dynamic}
  & Baseline         & \pmval{81.73}{0.55} & \pmval{65.35}{0.94} & \pmval{59.94}{0.68} & \pmval{67.95}{0.98} & \pmval{67.40}{0.86} & \pmval{62.43}{1.38} & \pmval{73.24}{0.60} & \pmval{77.33}{0.56} & \pmval{54.70}{0.98} \\
  & Rot-$z$ 15\textdegree & \pmval{76.73}{0.30} & \pmval{42.75}{0.25} & \pmval{48.20}{0.54} & \pmval{62.47}{0.44} & \pmval{63.69}{1.21} & \pmval{56.33}{1.09} & \pmval{73.33}{0.76} & \pmval{74.81}{1.59} & \pmval{48.99}{1.12} \\
  & Rot-$z$ 40\textdegree & \pmval{63.31}{0.59} & \pmval{27.10}{1.25} & \pmval{29.91}{1.14} & \pmval{36.85}{0.88} & \pmval{42.32}{0.55} & \pmval{30.78}{0.54} & \pmval{62.04}{1.76} & \pmval{59.84}{7.78} & \pmval{45.11}{2.06} \\
\midrule
PTv3 \citep{wu2024point}
  & Baseline         & \pmval{90.50}{0.13} & \pmval{79.10}{0.35} & \pmval{78.29}{1.20} & \pmval{82.52}{0.69} & \pmval{82.71}{1.01} & \pmval{78.85}{0.92} & \pmval{84.38}{0.85} & \pmval{84.33}{0.82} & \pmval{57.90}{3.23} \\
  & Rot-$z$ 15\textdegree & \pmval{89.91}{0.14} & \pmval{78.21}{0.38} & \pmval{73.92}{2.58} & \pmval{81.46}{0.85} & \pmval{78.46}{3.01} & \pmval{77.55}{1.13} & \pmval{80.72}{1.35} & \pmval{82.08}{0.75} & \pmval{49.63}{6.24} \\
  & Rot-$z$ 40\textdegree & \pmval{85.08}{0.59} & \pmval{62.97}{3.74} & \pmval{48.00}{5.04} & \pmval{63.45}{2.53} & \pmval{43.74}{5.24} & \pmval{42.64}{3.76} & \pmval{45.65}{4.17} & \pmval{54.10}{1.33} & \pmval{18.30}{8.52} \\
\midrule
Fast-TGCN \citep{li2024fine}
  & Baseline         & \pmval{94.80}{0.66} & \pmval{83.96}{0.93} & \pmval{77.78}{2.07} & \pmval{84.70}{0.42} & \pmval{77.58}{1.79} & \pmval{80.20}{1.06} & \pmval{82.09}{0.93} & \pmval{84.80}{0.87} & \pmval{56.12}{3.26} \\
  & Rot-$z$ 15\textdegree & \pmval{94.00}{0.78} & \pmval{81.52}{1.66} & \pmval{72.83}{1.80} & \pmval{79.39}{0.61} & \pmval{71.63}{2.85} & \pmval{75.02}{1.87} & \pmval{79.96}{1.22} & \pmval{82.70}{1.10} & \pmval{52.54}{1.79} \\
  & Rot-$z$ 40\textdegree & \pmval{84.90}{2.32} & \pmval{54.67}{3.50} & \pmval{38.50}{3.15} & \pmval{46.09}{2.71} & \pmval{42.71}{1.42} & \pmval{49.19}{0.76} & \pmval{63.91}{2.41} & \pmval{67.19}{2.51} & \pmval{43.16}{4.78} \\
\midrule
\rowcolor{gray!10} EAMS (ours)     & Baseline & \pmval{93.15}{0.75} & \pmval{66.18}{3.22} & \pmval{69.83}{1.93} & \pmval{73.80}{2.51} & \pmval{71.46}{2.22} & \pmval{67.08}{2.06} & \pmval{76.23}{1.75} & \pmval{74.93}{2.71} & \pmval{26.04}{5.75} \\
\rowcolor{gray!10} SRA+EAMS (ours) & Baseline & \pmval{94.29}{0.38} & \pmval{77.19}{2.03} & \pmval{75.33}{1.07} & \pmval{78.78}{1.28} & \pmval{72.40}{1.86} & \pmval{71.99}{1.50} & \pmval{75.01}{1.40} & \pmval{77.00}{1.65} & \pmval{19.52}{4.43} \\
\rowcolor{gray!10} VN+EAMS (ours)  & Baseline & \pmval{93.87}{0.71} & \pmval{80.16}{1.35} & \pmval{80.00}{0.68} & \pmval{84.85}{0.90} & \pmval{77.94}{0.91} & \pmval{78.20}{0.82} & \pmval{82.90}{0.71} & \pmval{82.17}{1.36} & \pmval{44.90}{7.71} \\
\bottomrule
\end{tabular}%
}
\end{table*}

\begin{figure*}[t]
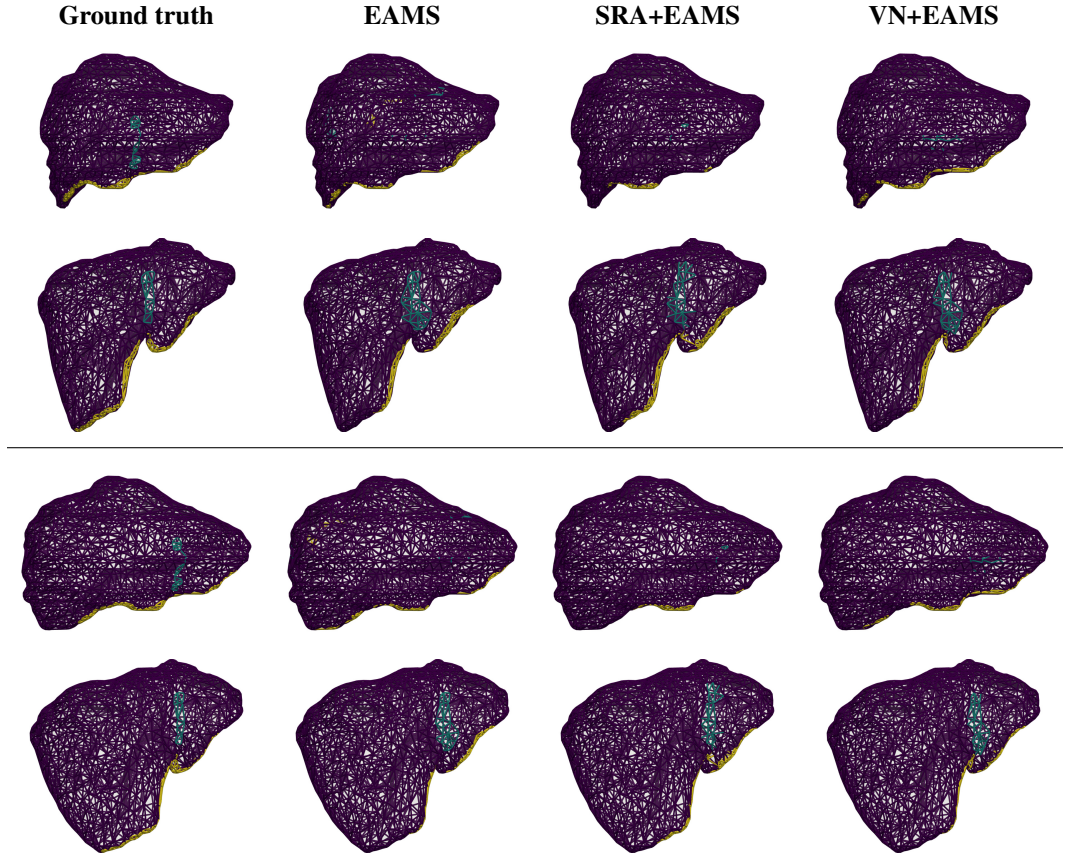

\centering
\makebox[0.245\textwidth][c]{\textbf{Ground truth}}\hfill
\makebox[0.245\textwidth][c]{\textbf{EAMS}}\hfill
\makebox[0.245\textwidth][c]{\textbf{SRA+EAMS}}\hfill
\makebox[0.245\textwidth][c]{\textbf{VN+EAMS}}

\par\smallskip

\begin{subfigure}[t]{0.245\textwidth}
  \qualimgliver{liver/emnn/base/LiTS-52_face_labels_xz_gt}
\end{subfigure}\hfill
\begin{subfigure}[t]{0.245\textwidth}
  \qualimgliver{liver/emnn_final/base/LiTS-52_face_labels_xz_pred}
\end{subfigure}\hfill
\begin{subfigure}[t]{0.245\textwidth}
  \qualimgliver{liver/sra_emnn_final/base/LiTS-52_face_labels_xz_pred}
\end{subfigure}\hfill
\begin{subfigure}[t]{0.245\textwidth}
  \qualimgliver{liver/vn_emnn_final/base/LiTS-52_face_labels_xz_pred}
\end{subfigure}

\par\smallskip
\begin{subfigure}[t]{0.245\textwidth}
  \qualimgliver{liver/emnn/base/LiTS-67_face_labels_xz_gt}
\end{subfigure}\hfill
\begin{subfigure}[t]{0.245\textwidth}
  \qualimgliver{liver/emnn_final/base/LiTS-67_face_labels_xz_pred}
\end{subfigure}\hfill
\begin{subfigure}[t]{0.245\textwidth}
  \qualimgliver{liver/sra_emnn_final/base/LiTS-67_face_labels_xz_pred}
\end{subfigure}\hfill
\begin{subfigure}[t]{0.245\textwidth}
  \qualimgliver{liver/vn_emnn_final/base/LiTS-67_face_labels_xz_pred}
\end{subfigure}

\par\medskip
\hrule
\par\smallskip

\begin{subfigure}[t]{0.245\textwidth}
  \qualimgliver{liver/emnn/rotate_40/LiTS-52_face_labels_xz_gt}
\end{subfigure}\hfill
\begin{subfigure}[t]{0.245\textwidth}
  \qualimgliver{liver/emnn_final/rotate_40/LiTS-52_face_labels_xz_pred}
\end{subfigure}\hfill
\begin{subfigure}[t]{0.245\textwidth}
  \qualimgliver{liver/sra_emnn_final/rotate_40/LiTS-52_face_labels_xz_pred}
\end{subfigure}\hfill
\begin{subfigure}[t]{0.245\textwidth}
  \qualimgliver{liver/vn_emnn_final/rotate_40/LiTS-52_face_labels_xz_pred}
\end{subfigure}

\par\smallskip
\begin{subfigure}[t]{0.245\textwidth}
  \qualimgliver{liver/emnn/rotate_40/LiTS-67_face_labels_xz_gt}
\end{subfigure}\hfill
\begin{subfigure}[t]{0.245\textwidth}
  \qualimgliver{liver/emnn_final/rotate_40/LiTS-67_face_labels_xz_pred}
\end{subfigure}\hfill
\begin{subfigure}[t]{0.245\textwidth}
  \qualimgliver{liver/sra_emnn_final/rotate_40/LiTS-67_face_labels_xz_pred}
\end{subfigure}\hfill
\begin{subfigure}[t]{0.245\textwidth}
  \qualimgliver{liver/vn_emnn_final/rotate_40/LiTS-67_face_labels_xz_pred}
\end{subfigure}

\caption{%
  Additional qualitative liver comparison for the EAMS-family methods on two representative cases from Figure~\ref{fig:liver-qual}, with the top half in the canonical orientation and the bottom half after a 40\textdegree{} $z$-axis rotation.
}
\label{fig:liver-qual-ours}
\end{figure*}

\begin{table*}[t]
\caption{%
  Per-class IoU (\%) for the right jaw half on 3D-IOSSeg \citep{li2024fine} (T21/41--T28/48).
  Results are mean $\pm$ std over five cross-validation folds under matched training augmentation, as in Table~\ref{tab:iosseg-main}. EAMS-family rotation variation is discussed in Section~\ref{sec:frame-selection}.
}
\label{tab:iosseg-pertooth-right}
\tablestyle{4pt}{1.1}
\adjustbox{max width=\textwidth}{%
\begin{tabular}{llcccccccc}
\toprule
Method & Condition
  & T21/41 $\uparrow$ & T22/42 $\uparrow$ & T23/43 $\uparrow$ & T24/44 $\uparrow$
  & T25/45 $\uparrow$ & T26/46 $\uparrow$ & T27/47 $\uparrow$ & T28/48 $\uparrow$ \\
\midrule
DGCNN \citep{wang2019dynamic}
  & Baseline         & \pmval{63.43}{0.90} & \pmval{57.66}{3.04} & \pmval{65.75}{2.90} & \pmval{65.83}{0.70} & \pmval{60.67}{0.21} & \pmval{69.37}{0.43} & \pmval{67.26}{0.87} & \pmval{56.41}{1.08} \\
  & Rot-$z$ 15\textdegree & \pmval{40.63}{3.40} & \pmval{37.18}{2.69} & \pmval{48.00}{1.62} & \pmval{58.42}{1.36} & \pmval{59.37}{0.24} & \pmval{68.05}{0.51} & \pmval{63.03}{1.08} & \pmval{62.32}{1.56} \\
  & Rot-$z$ 40\textdegree & \pmval{19.74}{0.37} & \pmval{7.18}{2.32} & \pmval{6.89}{3.99} & \pmval{20.04}{7.04} & \pmval{39.45}{3.13} & \pmval{55.41}{0.97} & \pmval{45.30}{1.28} & \pmval{55.88}{0.78} \\
\midrule
PTv3 \citep{wu2024point}
  & Baseline         & \pmval{80.91}{0.55} & \pmval{77.23}{0.84} & \pmval{82.48}{0.89} & \pmval{81.03}{1.35} & \pmval{77.92}{1.22} & \pmval{81.05}{1.91} & \pmval{80.43}{1.31} & \pmval{65.74}{2.51} \\
  & Rot-$z$ 15\textdegree & \pmval{75.90}{0.91} & \pmval{71.14}{1.30} & \pmval{76.78}{1.20} & \pmval{77.12}{1.89} & \pmval{75.29}{1.94} & \pmval{79.64}{1.03} & \pmval{81.89}{1.17} & \pmval{64.87}{8.63} \\
  & Rot-$z$ 40\textdegree & \pmval{50.94}{5.99} & \pmval{42.60}{4.47} & \pmval{47.51}{4.53} & \pmval{49.04}{5.62} & \pmval{39.88}{7.81} & \pmval{44.50}{6.34} & \pmval{48.70}{5.64} & \pmval{34.97}{4.79} \\
\midrule
Fast-TGCN \citep{li2024fine}
  & Baseline         & \pmval{85.53}{0.88} & \pmval{82.43}{1.41} & \pmval{87.28}{1.02} & \pmval{82.70}{0.68} & \pmval{74.43}{1.67} & \pmval{76.35}{0.73} & \pmval{78.14}{2.28} & \pmval{46.50}{9.52} \\
  & Rot-$z$ 15\textdegree & \pmval{81.77}{1.42} & \pmval{76.38}{2.04} & \pmval{77.91}{0.76} & \pmval{71.81}{1.39} & \pmval{66.08}{1.69} & \pmval{75.08}{1.23} & \pmval{78.09}{2.97} & \pmval{53.29}{6.24} \\
  & Rot-$z$ 40\textdegree & \pmval{57.53}{3.55} & \pmval{45.88}{4.40} & \pmval{41.52}{2.48} & \pmval{38.50}{1.54} & \pmval{41.32}{3.51} & \pmval{57.72}{2.94} & \pmval{64.30}{3.02} & \pmval{49.42}{6.71} \\
\midrule
\rowcolor{gray!10} EAMS (ours)     & Baseline & \pmval{68.50}{3.20} & \pmval{67.63}{2.31} & \pmval{73.54}{2.04} & \pmval{72.32}{1.89} & \pmval{67.50}{1.59} & \pmval{72.28}{1.31} & \pmval{73.61}{1.96} & \pmval{36.78}{3.26} \\
\rowcolor{gray!10} SRA+EAMS (ours) & Baseline & \pmval{79.57}{1.77} & \pmval{77.09}{1.69} & \pmval{77.78}{1.57} & \pmval{76.06}{1.38} & \pmval{68.19}{1.88} & \pmval{71.39}{1.84} & \pmval{77.80}{2.29} & \pmval{27.73}{8.92} \\
\rowcolor{gray!10} VN+EAMS (ours)  & Baseline & \pmval{80.87}{2.94} & \pmval{78.84}{2.25} & \pmval{82.66}{1.08} & \pmval{82.39}{1.28} & \pmval{75.29}{1.04} & \pmval{77.55}{1.52} & \pmval{77.68}{0.71} & \pmval{43.31}{5.77} \\
\bottomrule
\end{tabular}%
}
\end{table*}

\begin{table*}[t]
\caption{%
  Per-class IoU (\%) for the left jaw half on Teeth3DS \citep{ben20233dteethseg} (Gingiva and T11/31--T18/38).
  Single-fold results; no standard deviation reported.
  All conditions are reported empirically; frame-selection effects are explained in Section~\ref{sec:frame-selection}.
}
\label{tab:teeth3ds-pertooth-left}
\tablestyle{4pt}{1.1}
\adjustbox{max width=\textwidth}{%
\begin{tabular}{llccccccccc}
\toprule
Method & Condition
  & Gingiva $\uparrow$ & T11/31 $\uparrow$ & T12/32 $\uparrow$ & T13/33 $\uparrow$
  & T14/34 $\uparrow$ & T15/35 $\uparrow$ & T16/36 $\uparrow$ & T17/37 $\uparrow$ & T18/38 $\uparrow$ \\
\midrule
Fast-TGCN \citep{li2024fine}
  & Baseline              & 90.63 & 85.73 & 84.02 & 82.55 & 87.25 & 85.90 & 85.55 & 75.35 & 55.69 \\
  & Rot-$z$ 15\textdegree & 90.37 & 83.27 & 81.32 & 79.72 & 84.16 & 82.17 & 81.73 & 71.64 & 36.88 \\
  & Rot-$z$ 40\textdegree & 86.20 & 57.21 & 55.35 & 56.87 & 50.97 & 53.54 & 65.15 & 55.06 & 19.28 \\
\midrule
\rowcolor{gray!10} EAMS (ours)
  & Baseline              & 90.53 & 80.80 & 80.55 & 82.28 & 82.41 & 78.22 & 79.81 & 75.27 & 62.87 \\
\rowcolor{gray!10}
  & Rot-$z$ 15\textdegree & 90.68 & 81.17 & 81.33 & 82.97 & 82.98 & 78.55 & 79.86 & 75.48 & 70.10 \\
\rowcolor{gray!10}
  & Rot-$z$ 40\textdegree & 90.73 & 81.48 & 81.74 & 83.37 & 83.39 & 78.75 & 79.88 & 75.52 & 70.11 \\
\midrule
\rowcolor{gray!10} SRA+EAMS (ours)
  & Baseline              & 90.42 & 84.34 & 83.17 & 83.78 & 85.41 & 82.39 & 82.15 & 73.64 & 57.64 \\
\rowcolor{gray!10}
  & Rot-$z$ 15\textdegree & 90.74 & 84.32 & 83.58 & 84.50 & 85.67 & 82.65 & 83.01 & 75.86 & 69.54 \\
\rowcolor{gray!10}
  & Rot-$z$ 40\textdegree & 90.76 & 84.42 & 83.58 & 84.46 & 85.69 & 82.70 & 83.09 & 75.89 & 69.51 \\
\midrule
\rowcolor{gray!10} VN+EAMS (ours)
  & Baseline              & 91.18 & 85.11 & 84.07 & 83.88 & 84.99 & 81.79 & 82.04 & 76.32 & 36.04 \\
\rowcolor{gray!10}
  & Rot-$z$ 15\textdegree & 91.26 & 84.89 & 83.79 & 83.52 & 85.03 & 81.73 & 82.19 & 76.93 & 48.14 \\
\rowcolor{gray!10}
  & Rot-$z$ 40\textdegree & 91.28 & 84.99 & 83.95 & 83.75 & 85.15 & 81.86 & 82.28 & 77.05 & 48.14 \\
\bottomrule
\end{tabular}%
}
\end{table*}

\begin{table*}[t]
\caption{%
  Per-class IoU (\%) for the right jaw half on Teeth3DS \citep{ben20233dteethseg} (T21/41--T28/48).
  Single-fold results; no standard deviation reported.
  All conditions are reported empirically; frame-selection effects are explained in Section~\ref{sec:frame-selection}.
}
\label{tab:teeth3ds-pertooth-right}
\tablestyle{4pt}{1.1}
\adjustbox{max width=\textwidth}{%
\begin{tabular}{llcccccccc}
\toprule
Method & Condition
  & T21/41 $\uparrow$ & T22/42 $\uparrow$ & T23/43 $\uparrow$ & T24/44 $\uparrow$
  & T25/45 $\uparrow$ & T26/46 $\uparrow$ & T27/47 $\uparrow$ & T28/48 $\uparrow$ \\
\midrule
Fast-TGCN \citep{li2024fine}
  & Baseline              & 84.62 & 84.04 & 83.12 & 87.66 & 87.66 & 85.78 & 75.22 & 57.28 \\
  & Rot-$z$ 15\textdegree & 83.68 & 81.76 & 82.40 & 86.02 & 83.74 & 82.83 & 73.26 & 61.85 \\
  & Rot-$z$ 40\textdegree & 61.81 & 54.16 & 56.86 & 66.17 & 54.32 & 49.73 & 57.37 & 46.00 \\
\midrule
\rowcolor{gray!10} EAMS (ours)
  & Baseline              & 80.04 & 79.40 & 81.66 & 81.49 & 80.33 & 81.83 & 75.27 & 63.04 \\
\rowcolor{gray!10}
  & Rot-$z$ 15\textdegree & 80.17 & 79.51 & 81.64 & 81.96 & 81.15 & 82.53 & 75.73 & 68.62 \\
\rowcolor{gray!10}
  & Rot-$z$ 40\textdegree & 80.41 & 79.69 & 81.70 & 81.96 & 81.10 & 82.55 & 75.83 & 68.58 \\
\midrule
\rowcolor{gray!10} SRA+EAMS (ours)
  & Baseline              & 83.36 & 82.53 & 83.37 & 84.41 & 82.46 & 81.26 & 72.71 & 58.28 \\
\rowcolor{gray!10}
  & Rot-$z$ 15\textdegree & 83.67 & 83.02 & 83.75 & 84.86 & 82.92 & 82.38 & 74.76 & 68.98 \\
\rowcolor{gray!10}
  & Rot-$z$ 40\textdegree & 83.78 & 83.07 & 83.67 & 84.87 & 82.93 & 82.59 & 75.32 & 69.00 \\
\midrule
\rowcolor{gray!10} VN+EAMS (ours)
  & Baseline              & 85.05 & 84.63 & 84.72 & 84.84 & 82.59 & 82.01 & 75.19 & 44.90 \\
\rowcolor{gray!10}
  & Rot-$z$ 15\textdegree & 84.86 & 84.78 & 85.00 & 85.07 & 82.84 & 82.41 & 76.28 & 57.18 \\
\rowcolor{gray!10}
  & Rot-$z$ 40\textdegree & 85.02 & 84.80 & 85.00 & 85.06 & 82.81 & 82.41 & 76.18 & 57.16 \\
\bottomrule
\end{tabular}%
}
\end{table*}

\subsection{Liver surface segmentation: full robustness and distance results}

Table~\ref{tab:liver-robust-full} extends Table~\ref{tab:liver-main} with IoU and Hausdorff distance (HD), and reports Chamfer distance after multiplying by $100$ for readability, across all listed test-time conditions.
The EAMS-family rows show only minor variation across rotations, in contrast to the large degradation of the non-equivariant baselines.
Among the EAMS variants, SRA+EAMS gives the strongest ligament distance metrics, while VN+EAMS gives the strongest ridge distance metrics.

\begin{table*}[t]
\caption{%
  Full robustness results for liver surface segmentation on the dataset of \citep{zhang2025nested}.
  Results are mean $\pm$ std over five cross-validation folds.
  CD is reported after multiplying by $100$.
  Table~\ref{tab:liver-main} shows the corresponding Dice and CD summaries for the baseline and 40\textdegree{} conditions.
}
\label{tab:liver-robust-full}
\tablestyle{5pt}{1.15}
\adjustbox{max width=\textwidth}{%
\begin{tabular}{llcccccc}
\toprule
& & \multicolumn{3}{c}{Ligament} & \multicolumn{3}{c}{Ridge} \\
\cmidrule(lr){3-5}\cmidrule(lr){6-8}
Method & Condition
  & IoU (\%) $\uparrow$ & CD$\times 100$ $\downarrow$ & HD $\downarrow$
  & IoU (\%) $\uparrow$ & CD$\times 100$ $\downarrow$ & HD $\downarrow$ \\
\midrule
PointNet++ \citep{qi2017pointnet++}
  & Canonical & \pmval{24.83}{1.96} & \pmval{0.220}{0.030} & \pmval{0.0584}{0.0063} & \pmval{43.12}{0.73} & \pmval{0.099}{0.016} & \pmval{0.0604}{0.0020} \\
  & Rot-$z$ 15\textdegree & \pmval{16.94}{2.77} & \pmval{0.367}{0.090} & \pmval{0.0693}{0.0073} & \pmval{42.45}{1.80} & \pmval{0.108}{0.018} & \pmval{0.0649}{0.0057} \\
  & Rot-$z$ 40\textdegree & \pmval{7.62}{1.42} & \pmval{2.105}{1.187} & \pmval{0.1351}{0.0172} & \pmval{32.54}{1.00} & \pmval{0.879}{0.132} & \pmval{0.1588}{0.0114} \\
\midrule
DGCNN \citep{wang2019dynamic}
  & Canonical & \pmval{21.39}{2.06} & \pmval{0.416}{0.028} & \pmval{0.0837}{0.0047} & \pmval{42.32}{2.43} & \pmval{0.105}{0.032} & \pmval{0.0598}{0.0061} \\
  & Rot-$z$ 15\textdegree & \pmval{17.29}{2.24} & \pmval{0.711}{0.245} & \pmval{0.0915}{0.0098} & \pmval{41.62}{1.68} & \pmval{0.156}{0.005} & \pmval{0.0784}{0.0039} \\
  & Rot-$z$ 40\textdegree & \pmval{4.34}{0.54} & \pmval{3.237}{0.797} & \pmval{0.1600}{0.0199} & \pmval{27.99}{0.65} & \pmval{1.457}{0.169} & \pmval{0.2049}{0.0072} \\
\midrule
MeshGraphCNN \citep{zhang2025nested}
  & Canonical & \pmval{30.84}{1.01} & \pmval{0.658}{0.030} & \pmval{0.1158}{0.0049} & \pmval{44.56}{0.36} & \pmval{0.245}{0.009} & \pmval{0.0715}{0.0012} \\
  & Rot-$z$ 15\textdegree & \pmval{19.74}{0.56} & \pmval{1.283}{0.039} & \pmval{0.1445}{0.0063} & \pmval{39.55}{0.43} & \pmval{0.461}{0.009} & \pmval{0.1100}{0.0020} \\
  & Rot-$z$ 40\textdegree & \pmval{3.52}{0.27} & \pmval{6.595}{0.135} & \pmval{0.2726}{0.0043} & \pmval{25.54}{0.15} & \pmval{2.439}{0.021} & \pmval{0.2889}{0.0022} \\
\midrule
\rowcolor{gray!10} EAMS (ours)
  & Baseline           & \pmval{13.02}{0.75} & \pmval{3.928}{0.668} & \pmval{0.1353}{0.0153} & \pmval{41.12}{0.96} & \pmval{0.873}{0.064} & \pmval{0.1510}{0.0046} \\
  & Rot-$z$ 15\textdegree  & \pmval{13.02}{0.75} & \pmval{4.118}{0.527} & \pmval{0.1376}{0.0129} & \pmval{41.12}{0.94} & \pmval{0.874}{0.065} & \pmval{0.1512}{0.0044} \\
  & Rot-$z$ 40\textdegree  & \pmval{13.03}{0.75} & \pmval{4.307}{0.193} & \pmval{0.1399}{0.0094} & \pmval{41.11}{0.96} & \pmval{0.876}{0.065} & \pmval{0.1511}{0.0044} \\
\midrule
\rowcolor{gray!10} SRA+EAMS (ours)
  & Baseline           & \pmval{17.10}{1.54} & \pmval{1.722}{0.651} & \pmval{0.0953}{0.0051} & \pmval{39.35}{0.63} & \pmval{1.071}{0.372} & \pmval{0.1416}{0.0267} \\
  & Rot-$z$ 15\textdegree  & \pmval{17.13}{1.52} & \pmval{1.722}{0.650} & \pmval{0.0954}{0.0051} & \pmval{39.35}{0.65} & \pmval{1.070}{0.371} & \pmval{0.1416}{0.0267} \\
  & Rot-$z$ 40\textdegree  & \pmval{17.10}{1.54} & \pmval{1.722}{0.650} & \pmval{0.0953}{0.0051} & \pmval{39.34}{0.66} & \pmval{1.069}{0.371} & \pmval{0.1416}{0.0267} \\
\midrule
\rowcolor{gray!10} VN+EAMS (ours)
  & Baseline           & \pmval{13.42}{0.76} & \pmval{2.013}{0.592} & \pmval{0.1161}{0.0260} & \pmval{42.74}{2.15} & \pmval{0.610}{0.160} & \pmval{0.1283}{0.0168} \\
  & Rot-$z$ 15\textdegree  & \pmval{13.43}{0.77} & \pmval{2.014}{0.593} & \pmval{0.1161}{0.0260} & \pmval{42.75}{2.16} & \pmval{0.608}{0.159} & \pmval{0.1282}{0.0166} \\
  & Rot-$z$ 40\textdegree  & \pmval{13.43}{0.77} & \pmval{2.012}{0.591} & \pmval{0.1161}{0.0260} & \pmval{42.75}{2.16} & \pmval{0.609}{0.159} & \pmval{0.1282}{0.0168} \\
\bottomrule
\end{tabular}%
}
\end{table*}

\subsection{Intrinsic-feature and global-node ablations}
\label{sec:feature-ablations}

Table~\ref{tab:intra-ablations} complements the additive ablations in Section~\ref{sec:experiments} by removing features or the shared global node from EAMS.
Removing dihedrals reduces aneurysm IoU from 79.90\% to 77.12\%; removing both dihedrals and HKS reduces it to 49.86\%.
Removing the global node also lowers IoU for both classes.
These class-specific results use a separate configuration from the aggregate IntrA component ablation in Table~\ref{tab:context-ablations}; comparisons are made within each table.

\subsection{Global-token configuration}
\label{sec:token-ablations}

Table~\ref{tab:token-ablations} complements the feature and context ablations in Section~\ref{sec:experiments} by varying the size of the global token bank.
The SRA region-count sweep changes aneurysm IoU by less than 1.2 percentage points across the tested counts.
VN is more sensitive: increasing the virtual-node count from 16 to 64 reduces aneurysm IoU from 83.05\% to 75.38\%.
These single-fold results show that larger global token banks do not necessarily improve segmentation.
The SRA geometry-setting ablation is retained separately from the auxiliary-loss comparison in Table~\ref{tab:context-ablations}.

\begin{table*}[tbp]
\begin{minipage}[t]{0.44\textwidth}
\vspace{0pt}
\captionof{table}{Single-fold IntrA feature and global-node ablations. IoU in \%; PA = parent vessel, AN = aneurysm.}
\label{tab:intra-ablations}
\tablestyle{3.5pt}{1.05}
\centering
\begin{tabular}{lcc}
\toprule
Config. & PA IoU $\uparrow$ & AN IoU $\uparrow$ \\
\midrule
EAMS & 93.39 & 79.90 \\
- no global node & 92.29 & 76.90 \\
- no dihedrals & 93.05 & 77.12 \\
- no dihed. \& HKS & 78.89 & 49.86 \\
\bottomrule
\end{tabular}
\end{minipage}
\hfill
\begin{minipage}[t]{0.54\textwidth}
\vspace{0pt}
\captionof{table}{Single-fold IntrA geometry/token sweep. IoU in \%; PA = parent vessel, AN = aneurysm. Separate from Table~\ref{tab:context-ablations}.}
\label{tab:token-ablations}
\tablestyle{3.5pt}{1.05}
\begin{tabular}{lcc}
\toprule
Config. & PA IoU $\uparrow$ & AN IoU $\uparrow$ \\
\midrule
SRA+EAMS ($K=32$) & 94.77 & 84.12 \\
- no reg. geom. & 94.06 & 83.22 \\
- $K=8$ & 94.42 & 83.52 \\
- $K=16$ & 94.55 & 83.59 \\
- $K=64$ & 94.14 & 82.95 \\
\midrule
VN+EAMS ($V=16$) & 95.02 & 83.05 \\
- $V=8$ & 94.67 & 81.62 \\
- $V=32$ & 93.31 & 80.40 \\
- $V=64$ & 92.01 & 75.38 \\
\bottomrule
\end{tabular}
\end{minipage}
\end{table*}

\subsection{Random-axis rotation stress test}
\label{sec:random-axis}

We sample a rotation axis uniformly on the sphere and an angle uniformly from $[-90^\circ,90^\circ]$, applying it at test time without retraining or adaptation and recomputing geometry-dependent frame features.
This is a coordinate-frame stress test, not an estimate of clinical acquisition frequencies.
Tables~\ref{tab:random-axis-intra}, \ref{tab:random-axis-dental}, and~\ref{tab:random-axis-liver} report matched-augmentation baseline results; canonical values are already given in Tables~\ref{tab:intra-main}, \ref{tab:iosseg-main}, \ref{tab:liver-main}, and~\ref{tab:liver-robust-full}.
On 3D-IOSSeg, average IoU falls from 65.92\% to 28.18\% for DGCNN, from 80.45\% to 45.05\% for PTv3, and from 81.05\% to 45.59\% for Fast-TGCN.
EAMS-family results show negligible variation from canonical performance and are omitted from these tables: IntrA retains E(3) invariance under the oriented-surface convention, whereas dental/liver stability is empirical and subject to the frame-selection conditions in Section~\ref{sec:frame-selection}.
Teeth3DS Fast-TGCN random-axis performance is unreported.

\begin{table*}[tbp]
\begin{minipage}[t]{0.56\textwidth}
\vspace{0pt}
\captionof{table}{IntrA random-axis baseline results. Mean $\pm$ std over five folds; Dice and IoU in \% ($\uparrow$). EAMS-family scores equal Table~\ref{tab:intra-main}.}
\label{tab:random-axis-intra}
\tablestyle{2pt}{1.1}
\begin{tabular}{lcccc}
\toprule
& \multicolumn{2}{c}{Parent vessel} & \multicolumn{2}{c}{Aneurysm} \\
\cmidrule(lr){2-3}\cmidrule(lr){4-5}
Method & Dice & IoU & Dice & IoU \\
\midrule
DGCNN & \pmval{95.21}{0.19} & \pmval{91.13}{0.34} & \pmval{79.62}{1.30} & \pmval{69.59}{1.24} \\
PTv3 & \pmval{94.03}{1.04} & \pmval{89.13}{1.81} & \pmval{76.93}{4.80} & \pmval{67.45}{5.37} \\
\bottomrule
\end{tabular}
\end{minipage}
\hfill
\begin{minipage}[t]{0.42\textwidth}
\vspace{0pt}
\captionof{table}{3D-IOSSeg random-axis baseline results. Mean $\pm$ std over five folds. Dental EAMS-family scores have negligible variation from Table~\ref{tab:iosseg-main}.}
\label{tab:random-axis-dental}
\tablestyle{4pt}{1.1}
\begin{tabular}{lc}
\toprule
Method & Average IoU (\%) $\uparrow$ \\
\midrule
DGCNN & \pmval{28.18}{0.33} \\
PTv3 & \pmval{45.05}{0.65} \\
Fast-TGCN & \pmval{45.59}{1.12} \\
\bottomrule
\end{tabular}
\end{minipage}
\end{table*}

\begin{table*}[tbp]
\caption{Liver random-axis baseline results, mean $\pm$ std over five folds. Dice and IoU in \%; CD multiplied by 100. EAMS-family scores have negligible variation from Tables~\ref{tab:liver-main} and~\ref{tab:liver-robust-full}, subject to the frame-selection conditions in Section~\ref{sec:frame-selection}.}
\label{tab:random-axis-liver}
\label{tab:random-axis-liver-distance}
\tablestyle{5pt}{1.1}
\begin{tabular}{llcccc}
\toprule
& & \multicolumn{2}{c}{Overlap} & \multicolumn{2}{c}{Distance} \\
\cmidrule(lr){3-4}\cmidrule(lr){5-6}
Method & Target & Dice $\uparrow$ & IoU $\uparrow$ & CD$\times100$ $\downarrow$ & HD $\downarrow$ \\
\midrule
PointNet++ & Ligament & \pmval{20.36}{2.45} & \pmval{13.67}{1.94} & \pmval{1.351}{0.501} & \pmval{0.0946}{0.0095} \\
 & Ridge & \pmval{51.03}{1.45} & \pmval{35.87}{1.30} & \pmval{1.232}{0.088} & \pmval{0.1069}{0.0092} \\
\midrule
DGCNN & Ligament & \pmval{20.27}{2.37} & \pmval{13.15}{1.85} & \pmval{2.320}{0.425} & \pmval{0.1322}{0.0104} \\
 & Ridge & \pmval{46.80}{3.74} & \pmval{32.41}{2.92} & \pmval{1.391}{0.128} & \pmval{0.1241}{0.0123} \\
\midrule
MeshGraphCNN & Ligament & \pmval{21.94}{0.71} & \pmval{14.22}{0.72} & \pmval{7.506}{0.171} & \pmval{0.2291}{0.0045} \\
 & Ridge & \pmval{41.89}{0.17} & \pmval{28.98}{0.21} & \pmval{4.559}{0.432} & \pmval{0.2123}{0.0042} \\
\bottomrule
\end{tabular}
\end{table*}

\subsection{Liver failure analysis}
\label{sec:liver-failures}

On held-out liver meshes, higher mean ligament curvature is associated with lower ligament Dice for both EAMS and VN+EAMS.
We compute Spearman correlations between per-mesh Dice and mean target curvature within each fold; the association is negative in every fold, with median correlations of $-0.556$ for EAMS and $-0.559$ for VN+EAMS.
This shared pattern identifies highly curved ligament regions as a remaining challenge even with global context.
These descriptive associations neither establish causality nor imply rotation-induced frame instability, which is evaluated separately in Appendix~\ref{sec:pca-sensitivity}.

\subsection{Utility of coordinate pose}
\label{sec:pose-diagnostic}

We append centred raw $(x,y,z)$ coordinates to EMNN's scalar node features, deliberately introducing dependence on coordinate pose.
Table~\ref{tab:pose-diagnostic} reports a single-fold diagnostic on canonical inputs.
Raw coordinates improve both liver targets and 3D-IOSSeg, but reduce both IntrA target scores.
This supports a task-dependent interpretation: consistent acquisition alignment can supply useful predictive information, whereas arbitrary coordinate pose can distract from vascular geometry.
The diagnostic motivates hybrid pose-aware/equivariant models where dataset alignment is informative; it does not isolate the sole cause of the canonical accuracy gap.

\begin{table}[t]
\caption{Single-fold raw-coordinate diagnostic. Dice and average IoU are percentages ($\uparrow$); $\Delta$ is in percentage points. No standard deviation is reported.}
\label{tab:pose-diagnostic}
\tablestyle{5pt}{1.1}
\adjustbox{max width=\textwidth}{%
\begin{tabular}{lllrrr}
\toprule
Dataset & Metric & Target & EMNN & + raw XYZ & $\Delta$ \\
\midrule
Liver & Dice & Ligament & 19.33 & 21.29 & +1.96 \\
 & Dice & Ridge & 55.39 & 60.88 & +5.49 \\
IntrA & Dice & Parent & 96.43 & 94.57 & $-1.86$ \\
 & Dice & Aneurysm & 88.28 & 82.88 & $-5.40$ \\
3D-IOSSeg & IoU & Average & 71.15 & 75.94 & +4.79 \\
\bottomrule
\end{tabular}%
}
\end{table}

\subsection{PCA-coordinate sensitivity}
\label{sec:pca-sensitivity}

We test the frame-selection conditions in Section~\ref{sec:frame-selection} by applying 16 random-axis rotations per held-out mesh and recomputing the PCA coordinates used as input features.
For each mesh--rotation trial, we compare those coordinates and the resulting predictions with the unrotated reference, measuring maximum absolute coordinate change, predicted-label flip fraction, and maximum absolute class-probability change.
Table~\ref{tab:pca-sensitivity} reports median and 90th percentile across trials; flip fractions are not percentages.

\begin{table*}[t]
\caption{Sensitivity of the PCA coordinates used as input features and the resulting predictions. Each cell is median / 90th percentile across mesh--rotation trials. The dental evaluation uses Teeth3DS. Values measure coordinate change, label-flip fraction, and probability change, respectively; lower is better.}
\label{tab:pca-sensitivity}
\tablestyle{5pt}{1.1}
\adjustbox{max width=\textwidth}{%
\begin{tabular}{lccc}
\toprule
Dataset / frame & Coordinate change $\downarrow$ & Label flips $\downarrow$ & Probability change $\downarrow$ \\
\midrule
Liver / area-weighted PCA & $1.33\!\times\!10^{-6}$ / $7.70\!\times\!10^{-6}$ & 0 / 0 & $2.35\!\times\!10^{-6}$ / $2.47\!\times\!10^{-4}$ \\
Teeth3DS / dental PCA & $3.05\!\times\!10^{-5}$ / 54.3 & 0 / 0.133 & $3.60\!\times\!10^{-5}$ / 0.998 \\
\bottomrule
\end{tabular}%
}
\end{table*}

The area-weighted liver construction resolves two axis signs using weighted asymmetry/third moments and deterministic fallbacks, then obtains the third axis by a cross product.
All 960 evaluated liver trials have zero label flips, with coordinate changes at floating-point scale.
Dental median coordinate and prediction changes are also small.
Its upper-tail variation is consistent with the remaining $(X,Z)\mapsto(-X,-Z)$ ambiguity: orienting $Y$ by arch concavity and enforcing right-handedness does not select between these two frames.
For the representative Teeth3DS mesh \texttt{UHU2XH25\_upper}, a 54.3 maximum change concentrated in the cylindrical $z$ coordinate accompanies a 1.84\% label-flip rate and 0.901 maximum probability change.
This diagnostic complements the small overlap-score changes in the fixed-angle benchmark; it motivates a more complete anatomical sign rule without changing the interpretation of those measured benchmark results.
Exactly symmetric or degenerate shapes remain a separate frame-selection condition for any PCA construction.


\newpage
\section*{NeurIPS Paper Checklist}

\begin{enumerate}

\item {\bf Claims}
    \item[] Question: Do the main claims made in the abstract and introduction accurately reflect the paper's contributions and scope?
    \item[] Answer: \answerYes{}
    \item[] Justification: The abstract and introduction present the paper as a lightweight equivariant framework for robust anatomical mesh segmentation, and those claims match the task-dependent trade-offs reported in Sections~\ref{sec:intro}, \ref{sec:experiments}, and \ref{sec:discussion}.

\item {\bf Limitations}
    \item[] Question: Does the paper discuss the limitations of the work performed by the authors?
    \item[] Answer: \answerYes{}
    \item[] Justification: Section~\ref{sec:discussion} discusses the PCA frame-selection conditions, runtime overhead, the single-fold scope of targeted diagnostics, and the orientation-prior trade-off observed on liver segmentation; Appendix~\ref{sec:pca-sensitivity} reports PCA-coordinate and prediction sensitivity.

\item {\bf Theory assumptions and proofs}
    \item[] Question: For each theoretical result, does the paper provide the full set of assumptions and a complete (and correct) proof?
    \item[] Answer: \answerNA{}
    \item[] Justification: The paper is an empirical study of equivariant architectures for anatomical mesh segmentation and does not present new theorems or formal proofs.

    \item {\bf Experimental result reproducibility}
    \item[] Question: Does the paper fully disclose all the information needed to reproduce the main experimental results of the paper to the extent that it affects the main claims and/or conclusions of the paper (regardless of whether the code and data are provided or not)?
    \item[] Answer: \answerYes{}
    \item[] Justification: The main paper and appendix describe the datasets, preprocessing pipeline, feature construction, model variants, losses, hyperparameters, and evaluation protocols needed to reproduce the reported experiments; see Sections~\ref{sec:methods}--\ref{sec:experiments} and Appendix~\ref{sec:app-data}--\ref{sec:app-training}.

\item {\bf Open access to data and code}
    \item[] Question: Does the paper provide open access to the data and code, with sufficient instructions to faithfully reproduce the main experimental results, as described in supplemental material?
    \item[] Answer: \answerYes{}
    \item[] Justification: The experiments use public datasets. Training code, preprocessing, split definitions, and example configurations are publicly available at \coderepourl; see Appendix~\ref{sec:app-training}.

\item {\bf Experimental setting/details}
    \item[] Question: Does the paper specify all the training and test details (e.g., data splits, hyperparameters, how they were chosen, type of optimizer) necessary to understand the results?
    \item[] Answer: \answerYes{}
    \item[] Justification: The paper specifies the data splits and metrics in Section~\ref{sec:experiments}, and the appendix provides the optimizer, schedule, batch sizes, loss weights, architecture dimensions, preprocessing, and augmentation details in Appendix~\ref{sec:app-training}.

\item {\bf Experiment statistical significance}
    \item[] Question: Does the paper report error bars suitably and correctly defined or other appropriate information about the statistical significance of the experiments?
    \item[] Answer: \answerYes{}
    \item[] Justification: For the cross-validated datasets, the paper reports mean $\pm$ standard deviation over five folds in the main result tables and states the single-fold Teeth3DS and targeted pose/context ablation protocols explicitly in Section~\ref{sec:experiments} and Appendix~\ref{sec:pose-diagnostic}.

\item {\bf Experiments compute resources}
    \item[] Question: For each experiment, does the paper provide sufficient information on the computer resources (type of compute workers, memory, time of execution) needed to reproduce the experiments?
    \item[] Answer: \answerYes{}
    \item[] Justification: Appendix~\ref{sec:app-training} reports that all experiments were run on a single H100 GPU and provides per-epoch runtimes for EAMS together with runtime multipliers for the augmented variants.
    
\item {\bf Code of ethics}
    \item[] Question: Does the research conducted in the paper conform, in every respect, with the NeurIPS Code of Ethics \url{https://neurips.cc/public/EthicsGuidelines}?
    \item[] Answer: \answerYes{}
    \item[] Justification: The work uses public medical-imaging datasets and methodological model development, and no aspect of the research process is known to conflict with the NeurIPS Code of Ethics.

\item {\bf Broader impacts}
    \item[] Question: Does the paper discuss both potential positive societal impacts and negative societal impacts of the work performed?
    \item[] Answer: \answerYes{}
    \item[] Justification: Appendix~\ref{sec:app-broader-impacts} discusses the intended clinical benefits of more robust anatomical segmentation and the main risk of harm from incorrect predictions in downstream clinical workflows.
    
\item {\bf Safeguards}
    \item[] Question: Does the paper describe safeguards that have been put in place for responsible release of data or models that have a high risk for misuse (e.g., pre-trained language models, image generators, or scraped datasets)?
    \item[] Answer: \answerNA{}
    \item[] Justification: The paper does not release high-risk generative models, scraped datasets, or other assets that would require special misuse safeguards.

\item {\bf Licenses for existing assets}
    \item[] Question: Are the creators or original owners of assets (e.g., code, data, models), used in the paper, properly credited and are the license and terms of use explicitly mentioned and properly respected?
    \item[] Answer: \answerYes{}
    \item[] Justification: The manuscript cites the original datasets and prior methods, and the appendix credits the reused FDI illustration.

\item {\bf New assets}
    \item[] Question: Are new assets introduced in the paper well documented and is the documentation provided alongside the assets?
    \item[] Answer: \answerNo{}
    \item[] Justification: The implementation is released at \coderepourl, with configurations described in Appendix~\ref{sec:app-training}; comprehensive asset documentation has not been verified.

\item {\bf Crowdsourcing and research with human subjects}
    \item[] Question: For crowdsourcing experiments and research with human subjects, does the paper include the full text of instructions given to participants and screenshots, if applicable, as well as details about compensation (if any)? 
    \item[] Answer: \answerNA{}
    \item[] Justification: The paper does not involve crowdsourcing or prospective data collection from human subjects.

\item {\bf Institutional review board (IRB) approvals or equivalent for research with human subjects}
    \item[] Question: Does the paper describe potential risks incurred by study participants, whether such risks were disclosed to the subjects, and whether Institutional Review Board (IRB) approvals (or an equivalent approval/review based on the requirements of your country or institution) were obtained?
    \item[] Answer: \answerNA{}
    \item[] Justification: The paper uses existing public datasets and does not report prospective human-subjects experimentation requiring participant instructions or new IRB review.

\item {\bf Declaration of LLM usage}
    \item[] Question: Does the paper describe the usage of LLMs if it is an important, original, or non-standard component of the core methods in this research? Note that if the LLM is used only for writing, editing, or formatting purposes and does \emph{not} impact the core methodology, scientific rigor, or originality of the research, declaration is not required.
    \item[] Answer: \answerNA{}
    \item[] Justification: LLMs are not part of the proposed segmentation models, training procedure, or experimental methodology.

\end{enumerate}

\end{document}